\documentclass[10pt,twocolumn,letterpaper]{article}
\PassOptionsToPackage{table}{xcolor}
\usepackage[iccvfinal]{iccv}
\input{preamble}

\usepackage[accsupp]{axessibility}
\usepackage{times}
\usepackage{epsfig}
\usepackage{graphicx}
\usepackage{amsmath}
\usepackage{amssymb}
\usepackage{enumitem}
\usepackage{color}

\definecolor{grannysmithapple}{rgb}{0.66, 0.89, 0.63}
\definecolor{goldenyellow}{rgb}{1.0, 0.87, 0.0}
\definecolor{lemonchiffon}{rgb}{1.0, 0.98, 0.8}
\definecolor{iccvblue}{rgb}{0.21,0.49,0.74}
\usepackage[pagebackref,breaklinks,colorlinks,allcolors=iccvblue]{hyperref}

\usepackage{algorithm2e}
\RestyleAlgo{ruled}
\SetKwComment{Comment}{/* }{ */}

\usepackage{float}

\usepackage{multirow}

\usepackage{caption}
\usepackage{booktabs}

\newcommand{\name}{\textsc{AV-DAR}\xspace}

\def\paperID{14964} % *** Enter the Paper ID here
\def\confName{ICCV}
\def\confYear{2025}

\title{Differentiable Room Acoustic Rendering with Multi-View Vision Priors}

\author{Derong Jin\\
University of Maryland, College Park\\
{\tt\small djin77@umd.edu}
\and
Ruohan Gao\\
University of Maryland, College Park\\
{\tt\small rhgao@umd.edu}
}

\begin{document}

\maketitle

%%%%%%%%% ABSTRACT
\begin{abstract}
An immersive acoustic experience enabled by spatial audio is just as crucial as the visual aspect in creating realistic virtual environments. 
However, existing methods for room impulse response estimation rely either on data-demanding learning-based models or computationally expensive physics-based modeling. In this work, we introduce Audio-Visual Differentiable Room Acoustic Rendering (\name), a framework that leverages visual cues extracted from multi-view images and acoustic beam tracing for physics-based room acoustic rendering.
Experiments across six real-world environments from two datasets demonstrate that our multimodal, physics-based approach is efficient, interpretable, and accurate, significantly outperforming a series of prior methods. Notably, on the Real Acoustic Field dataset, \name achieves comparable performance to models trained on 10 times more data while delivering  relative gains ranging from 16.6\% to 50.9\% when trained at the same scale.
Project Page: \href{https://humathe.github.io/avdar/}{https://humathe.github.io/avdar/}.
\end{abstract}

%%%%%%%%% BODY TEXT
\section{Introduction}
\begin{figure}[t]
    \centering
    \includegraphics[width=\linewidth, draft=false]{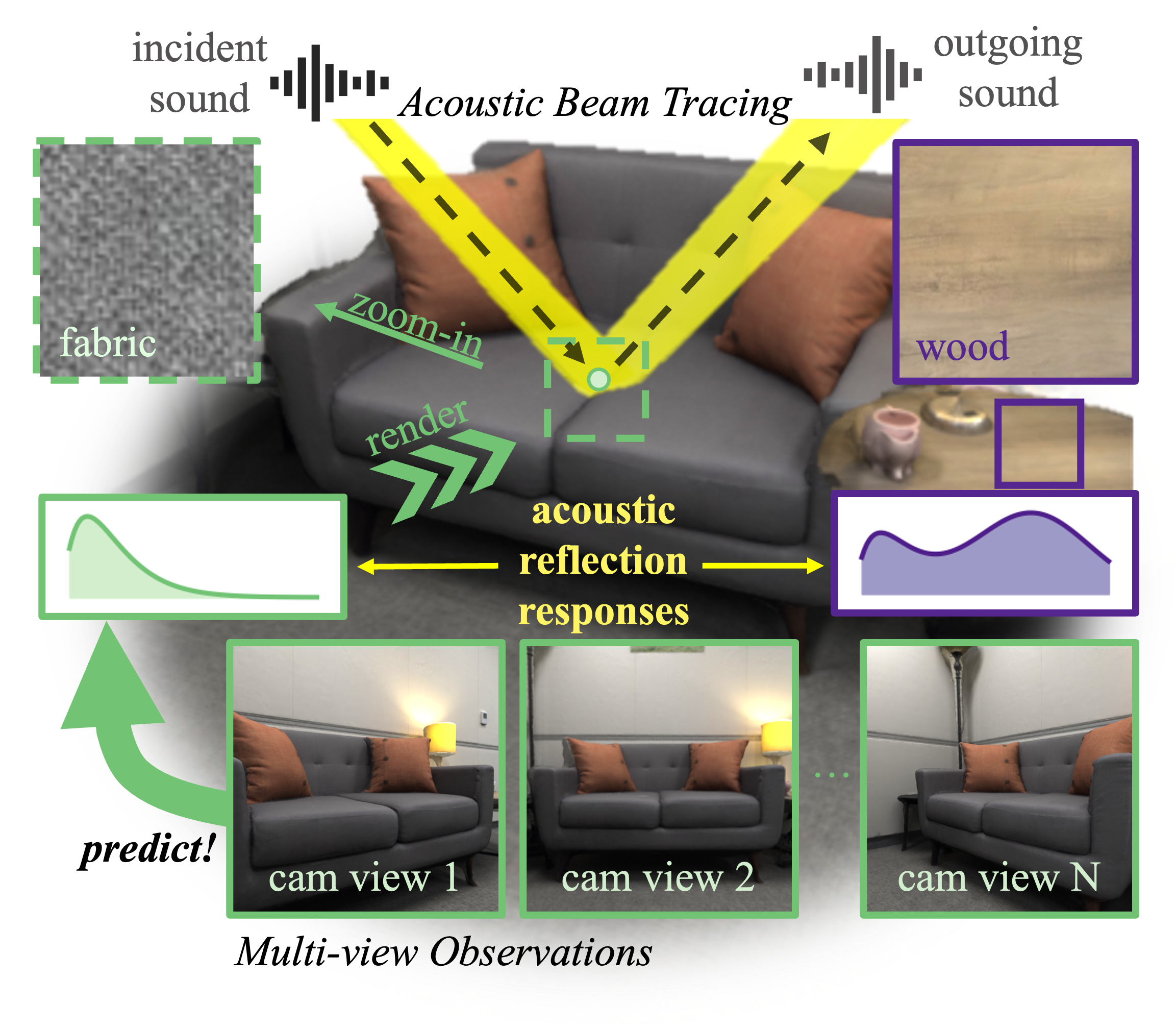}
    \vspace{-0.35in}
    \caption{Our differentiable room acoustic rendering framework combines multi-view visual observations and acoustic beam tracing for efficient and accurate room impulse response (RIR) prediction. By analyzing the visual cues of surfaces (\eg, fabric vs. wood), it infers acoustic reflection responses for accurately rendering RIRs through physics-based, end-to-end optimization. }
    \label{fig:teaser}
    \vspace{-0.1in}
\end{figure}

\label{sec:intro}

Spatial audio is a fundamental component of immersive multimedia experiences and is often regarded as “half the experience” in virtual and augmented reality (VR/AR) applications. For example, spatialized sound can help users navigate real environments~\cite{spatialnavigation}, 
and precise spatial positioning of audio sources enhances natural remote communication and engagement~\cite{oviedo2024social}.
Accurately modeling spatial audio unlocks a wide range of immersive applications, from entertainment and gaming to education and telepresence.

Recreating the spatial acoustic experience is analogous to novel-view synthesis \cite{martinbrualla2020nerfw,mildenhall2020nerf} in vision, where the goal is to learn from sparse images to synthesize photorealistic images from arbitrary viewpoints. Similarly, novel-view acoustic synthesis~\cite{avr,Liang2023NeuralAC,luo2022learning,su2022inras,hearinganythinganywhere2024} aims to render the received sound at any listener location within a scene. A widely used representation for this task is the \emph{Room Impulse Response} (RIR) \cite{rirrenderer,roomimpulseresponse}, which describes how an emitted sound propagates through space---including reflections and diffractions---before reaching a listener. Accurately estimating RIR for every possible source-listener pair would, in principle, enable realistic spatial acoustic rendering. 

Existing methods for estimating RIRs generally fall into two broad categories: \emph{learning-based} and \emph{physics-based}. Learning-based approaches \cite{liang23avnerf,luo2022learning,AVRIR,su2022inras,bhosale2024avgs,liu2025haae} treat RIR estimation as a regression task trained on densely measured ground-truth RIRs. Although effective for fitting simulated scenes and offering efficient inference, they can be difficult to deploy in real-world settings due to high data requirements and the lack of physically grounded guarantees. In contrast, physics-based approaches \cite{avr,hearinganythinganywhere2024} rely on explicit acoustic models such as the image-source method \cite{hearinganythinganywhere2024} or volume rendering \cite{avr}. However, both are impractical for large scenes with sparse training samples: the image-source method requires significant computation to enumerate reflection paths and becomes prohibitively expensive in complex environments, while volume rendering demands extensive 3D samples for RIR reconstruction and considerable training data to learn a neural acoustic field.

Our key insight is that while sound fundamentally differs from light---traveling more slowly and exhibiting time-of-arrival variations---both are influenced by the same geometry and surface material properties within a given scene. As shown in Fig.~\ref{fig:teaser}, a surface region's acoustic property often strongly correlates with visual appearance due to the same underlying materials. For example, smooth, hard surfaces like glass tend to reflect high-frequency sound, whereas rough, deformable materials such as carpets primarily absorb high-frequency components and reflect lower-frequency waves. Leveraging visual observations to estimate surface acoustic properties in a physics-based differentiable framework could potentially enable more accurate and efficient learning of room acoustic parameters.

To realize this intuition, we introduce an \emph{Audio-Visual Differentiable Room Acoustic Rendering} (\name) framework, which leverages visual cues to guide the differentiable rendering of RIRs in an end-to-end manner. By aligning multi-view image features from camera space to 3D scene space via a cross-attention mechanism, our approach decouples view-dependent visual information, forming a unified, material-aware representation in the scene space. This representation serves as a robust foundation for learning reflection properties and enables accurate RIR estimation at unobserved locations. Additionally, we employ beam tracing~\cite{beam1,beam2} to search for specular reflection paths and model the acoustic field, which requires fewer training samples than volume rendering and significantly reduces computation time compared to the image-source method. 

Experiments on six real-world environments~\cite{chen2024RAF,hearinganythinganywhere2024} show that our approach significantly outperforms both learning-based and physics-based baselines.
For example, on the Real Acoustic Field dataset~\cite{chen2024RAF}, our model achieves comparable performance to existing methods trained on roughly 10$\times$ RIR measurements while delivering 16.6\% to 50.9\% improvement when trained at the same scale. Moreover, qualitative analysis confirms that our method learns material-specific acoustic reflection responses that closely align with the visual characteristics of the scene.

Our main contributions are threefold: First, we propose a \emph{physics-based} \emph{differentiable} room acoustic rendering pipeline that not only learn from sparse, real-world RIR measurements but is also \emph{efficient}, \emph{interpretable}, and \emph{accurate}. Second, we are the first to integrate acoustic beam tracing within an  end-to-end differentiable framework, enabling efficient computation of reflection responses. Third, our approach leverages multi-view images to capture material-aware visual cues that correlate with acoustic reflection properties, achieving significantly more accurate RIR rendering than prior methods.

\section{Related Work}
\label{sec:relatedwork}

\noindent\textbf{Learning-Based RIR Prediction.} A growing body of work leverages machine learning techniques to approximate room impulse responses (RIRs) directly. Early methods often rely on large sets of measured or simulated RIRs to train neural networks capable of predicting RIRs at new positions. For instance, Ratnarajah \etal~\cite{fast-rir} use a generative adversarial network (GAN) to synthesize RIRs, while later efforts incorporate scene meshes~\cite{Mesh2Ir} or visual signals~\cite{AVRIR, liu2025haae} to condition RIR generation. Continuous implicit neural representations have also been explored to model high-fidelity acoustic fields within individual scenes~\cite{liang23avnerf,Liang2023NeuralAC,luo2022learning, deepir,su2022inras, bhosale2024avgs}. However, unlike our approach, which can learn from sparse RIR measurments, these methods often rely on dense measurements and/or simulated data, requiring up to 1,000 measured RIRs in order to accurately interpolate to new positions within the same environment~\cite{deepir}.

\vspace{0.05in}

\noindent\textbf{Physics-Based Room Acoustics Modeling.} Classical acoustic modeling generally follows one of two paradigms: \emph{wave-based} or \emph{geometric} approaches. Wave-based methods directly solve the wave equation to model how sound propagates through air and interacts with surfaces~\cite{wave3,wave1,wave5,wave4,wave2}. These methods accurately capture inference and diffraction but can become computationally prohibitive for large domains or higher frequencies. In contrast, geometric acoustics approximates sound propagation using rays or beams. Techniques such as the \emph{image-source method}~\cite{imgsrc} account for specular reflections by spawning virtual sources, but the number of virtual sources grows exponentially with reflection order~\cite{chen22soundspaces2}. Other geometric approaches, such as \emph{ray tracing}~\cite{ray1,ray2} and \emph{beam tracing}~\cite{beam1,beam2,beam4,beam3}, use stochastic sampling to achieve more efficient rendering and more accurate late reverberation modeling compared to image-source methods~\cite{ray3}. Our method is also physics-based and uses beam tracing, but we integrate it into our differentiable room acoustic rendering pipeline for end-to-end optimization.

\vspace{0.05in}

\noindent\textbf{Audio-Visual Room Acoustics Learning.} Recent inspiring work integrates visual and audio signals on an array of interesting tasks related to room acoustics, including predicting how a given audio signal would sound in a scene visually depicted in images and videos~\cite{learning-av-dereverb,adverb,VisualAM,img2reverb,li2024self}, converting monaural audio to spatial audio based on visual spatial cues from the environment~\cite{li2018scene,morgado2018self,garg2021geometry,garg2023visually,gao2019visual-sound}, learning image representations, scene structures, and human locations through echolocation and ambient sound~\cite{gao2020visualechoes,christensen2020batvision,chen2021structure,wang2023soundcam}, navigating in simulated and real room environments based on audio-visual observations~\cite{gao2023sonicverse,chen2021waypoints,gan2020look}, and synthesizing binaural sound from novel viewpoints~\cite{nvas3d, nvas, SLfM, liang23avnerf, bee, fewavlearningenvacoustics}. Different from all of them, we incorporate multi-view visual cues into a differentiable room acoustic rendering pipeline for accurate RIR prediction.

\vspace{0.05in}

\noindent\textbf{Differentiable Acoustic Rendering.} Differentiable rendering has become a powerful tool in graphics and  vision, enabling image-based training for a wide range of reconstruction tasks~\cite{graphics3,graphics2,graphics1}. A similar approach can be applied to acoustic rendering, allowing inverse problems to be solved by optimizing acoustic-related physical properties using gradient-based optimization~\cite{diffocean, diffimpact, ddsp, jin2024diffsound}. In room acoustics, DiffRIR~\cite{hearinganythinganywhere2024} introduces a differentiable image-source renderer that jointly learns source and scene properties, and AVR~\cite{avr} applies a differentiable volume renderer to train a neural acoustic field. Differently, we leverage beam tracing~\cite{beam1,beam2} to efficiently search for specular reflection paths, achieving higher accuracy than ray tracing and significantly reduced computation time compared to volume rendering or image-source methods~\cite{ray3}.

\section{Approach}
\subsection{Preliminaries}
\label{sec:problem-formulation}

\begin{figure*}[t]
    \centering
    \includegraphics[width=1\linewidth,draft=false]{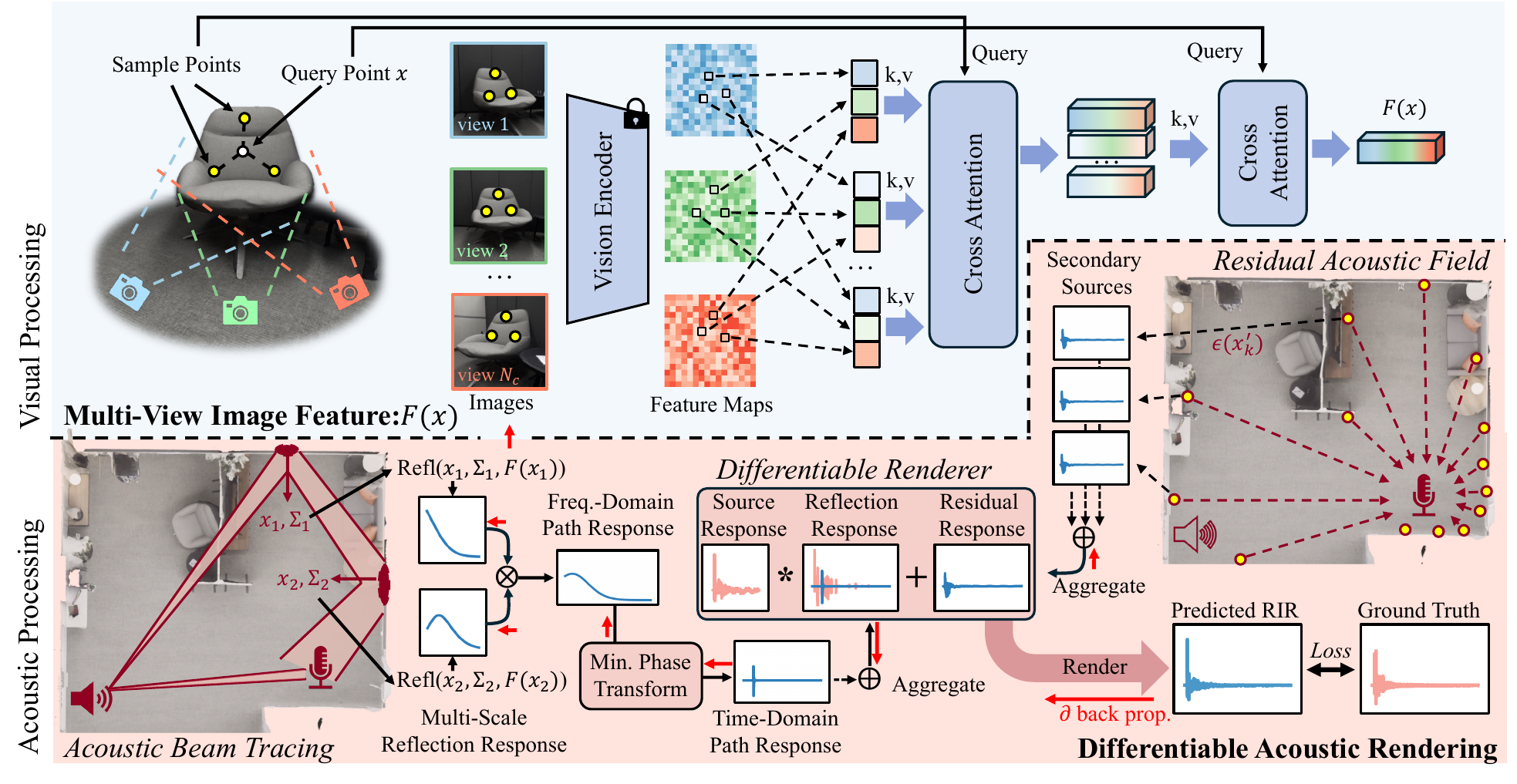}
    \vspace{-0.25in}
    \caption{
    \textbf{Method Overview.} Our framework contains two main components for rendering the room impulse response (RIR): \textbf{(1) Visual Processing (top)}: Multi-view images of the scene are passed through a pre-trained vision encoder
    to extract pixel-aligned features at sampled points on the room surface. We then apply cross-attention both across views for each sampled point and across sampled points of query ${\bf x}$ to obtain a unified, material-aware visual feature $\mathrm{F}(\mathbf{x})$ (detailed in Section~\ref{sec:vision}). \textbf{(2) Acoustic Processing (bottom):} On the left, we illustrate our acoustic beam tracing procedure (Section~\ref{sec:beamtracing}), where we sample specular paths and compute the path reflection response, conditioned on both the positional encoding (Section~\ref{sec:mip-rr}) and the visual feature $\mathrm{F}({\bf x})$. On the right, we show how we model the residual acoustic field (Section~\ref{sec:residual}) by treating every point on the surface as a secondary sound source and integrating its contribution via Monte-Carlo integration. The entire pipeline is fully differentiable, enabling end-to-end optimization of both acoustic and visual parameters.
    }
    \label{fig:pipeline}
\end{figure*}

Our goal is to learn a time-domain room impulse response function from sparse training data:
\vspace{-0.05in}
\begin{equation}
    \widehat{\mathrm{RIR}}(\mathbf{x}_a, \mathbf{x}_b, \mathbf{p}_a, t).
\vspace{-0.05in}
\end{equation}
Here, $\mathbf{x}_a$, $\mathbf{x}_b$, $\mathbf{p}_a$  denote the speaker location, the listener position, and the source orientation, respectively.

During training, besides using a sparse set of ground-truth RIR measurements, we leverage multi-view images to capture the scene's visual information. Formally, we assume a set of $N_c$ images with known camera intrinsics $\pi$ and extrinsics $P^{(i)}$:
\vspace{-0.05in}
\begin{equation}
    \bigl\{\{I^{(i)}, \pi, P^{(i)}\}{\Big|}\; i=1,\cdots, N_c\bigr\}.
\vspace{-0.05in}
\end{equation}
Our motivation of using multi-view images arises from the fact that, while RIR measurements encode the final results of acoustic wave propagation---including direct sound, early reflections, and late reverberations---they make it difficult to directly infer the underlying surface reflection properties. Multi-view images, on the other hand, capture visual material and geometric cues that are essential for predicting plausible reflection responses, effectively complementing the acoustic modality.

In practical applications, once $\widehat{\mathrm{RIR}}$ is learned, convolving it with an arbitrary input signal $h(t)$ yields the predicted audio at $\mathbf{x}_b$:
\vspace{-0.05in}
\begin{equation}
    h_b(t) = h(t) \; * \; \widehat{\mathrm{RIR}}(\mathbf{x}_a, \mathbf{x}_b, \mathbf{p}_a, t),
\vspace{-0.05in}
\end{equation}
thus enabling realistic spatial acoustic rendering. 

\subsection{Overview of the \textbf{\name} Framework}
\label{sec:overview}

We now introduce our framework, \name, which estimates room impulse responses (RIRs) at novel source-listener pair locations from sparse RIR measurements, multi-view images, and rough geometry of the room (\eg, expressed as a small number of planes). Following the decomposition in \cite{hearinganythinganywhere2024}, we model the RIR as:
\begin{equation}
    \widehat{\mathrm{RIR}}(t) = \sum_{\tau} s(\tau; \Theta_1) \cdot R(t-\tau; \Theta_2) + r(t; \Theta_3),
    \label{eq:rir-decomposition}
\vspace{-0.1in}
\end{equation}
where:
\begin{itemize}[leftmargin=*, noitemsep]
    \item \({s(t; \Theta_1)}\) is a learnable time-indexed vector representing the speaker’s source impulse response.
    \item \({R(t; \Theta_2)}\) is the integrated reflection response. We compute this term using beam tracing and differentiable rendering (Section~\ref{sec:beamtracing}), which is efficient compared to image-source methods and more accurate than ray tracing. To adapt to the beam tracing approach, we develop a multi-scale reflection response (Section~\ref{sec:mip-rr}). In addition, our multi-view vision module (Section~\ref{sec:vision}) extracts visual material cues to condition the reflection response prediction explicitly on surface properties.
    \item \({r(t; \Theta_3)}\) is the position-dependent residual impulse response that models high-order reflections, diffraction, and late reverberation (Section~\ref{sec:residual}).
\end{itemize}

Overall, \name integrates all components into an end-to-end differentiable pipeline, enabling gradient-based optimization for accurate RIR rendering.

\subsection{Acoustic Beam Tracing}
\label{sec:beamtracing}

Here we aim to develop a differentiable acoustic rendering model that supports gradient-based optimization of reflection responses. This requires an efficient method to compute the integrated reflection response while maintaining differentiability for backpropagation.

Traditional image-source methods~\cite{imgsrc} yield accurate specular reflections but are computationally prohibitive in large scenes due to their exponential complexity. Although ray tracing~\cite{ray1,ray3,ray2} offers faster, stochastic sampling of paths, it often fails to capture specular reflections when the source and listener are modeled as infinitesimal points. To overcome these limitations, we adopt beam tracing~\cite{beam1,beam2,beam3} as our acoustic primitive. In contrast to ray tracing—which treats sound as infinitesimally thin lines—beam tracing represents sound as volumetric, cone-shaped beams ({see Supp. for an illustration}); a listener is considered ``hit'' if it lies within a beam, ensuring robust detection of specular paths without relying on artificial volumetric approximations.

We uniformly sample $N_d$ beams from the source using a Fibonacci lattice to ensure even angular coverage and select a small apex angle to prevent beam overlap while maximizing spatial efficiency. Assuming narrow beams, we neglect beam splitting. The beam tracing function defines the set of valid reflection paths between source $\mathbf{x}_a$ and listener $\mathbf{x}_b$ as:
\vspace{-0.05in}
\begin{equation}
\mathcal{P}(\mathbf{x}_a, \mathbf{x}_b) = \{ \tilde{\mathbf{x}}_k \}_{k=1}^{N}.
\vspace{-0.05in}
\end{equation}
For each path $\tilde{\mathbf{x}}$, we compute the frequency response by combining the frequency-dependent reflection response at each hit point, $\mathrm{Refl}(\mathbf{x})[f]$, with the source's directional response, $\mathbf{D}_{\tilde{x}}$. The total frequency-domain attenuation is:
\vspace{-0.05in}
\begin{equation}
\prod_{\mathbf{x}_j \in \tilde{\mathbf{x}}} \mathrm{Refl}(\mathbf{x}_j)[f] \cdot \mathbf{D}_{\tilde{x}}[f].\label{eq:path-freq}
\vspace{-0.05in}
\end{equation}
This product is transformed into the time domain via a minimum phase transform~\cite{MPT}:
\begin{equation}
\kappa(\tilde{\mathbf{x}}, t) = \text{MinPhase}\left\{ \mathbf{D}_{\tilde{x}} \circ \prod_{\mathbf{x}_j \in \tilde{\mathbf{x}}} \mathrm{Refl}(\mathbf{x}_j) \right\}(t).
\end{equation}
In addition, we account for attenuation and delay due to air absorption and propagation loss using a propagation operator:
\begin{equation}
\mathcal{S}_{\tau}\{h\}(t) = \frac{\exp(-a_0 \tau)}{v_{\mathrm{sound}} \, \tau} \, h(t-\tau),
\end{equation}
where $\exp(-a_0 \tau)$ models air absorption, $(v_{\mathrm{sound}} \, \tau)^{-1}$ represents propagation loss, and $h(t-\tau)$ applies the appropriate delay. For a path with total travel time $\tilde{t}$, the final contribution to the reflection response is $\mathcal{S}_{\tilde{t}}\left\{ \kappa(\tilde{\mathbf{x}}, t) \right\}$.

Finally, the integrated reflection response is obtained by summing the contributions of all valid paths:
\vspace{-0.05in}
\begin{equation}
R(t) = \sum_{\tilde{\mathbf{x}}_k \in \mathcal{P}} \mathcal{S}_{\tilde{t}_k}\left\{ \kappa(\tilde{\mathbf{x}}_k, t) \right\}.
\end{equation}
Our differentiable formulation, combined with beam tracing, enables efficient optimization of reflection response learning and visual feature extraction, as discussed next.

\subsection{Multi-Scale Reflection Response}
\label{sec:mip-rr}

In a volumetric beam framework, the \emph{contact region} between a beam and a surface grows as the beam propagates farther from the source, causing the reflected energy to span a larger patch of the surface. This phenomenon naturally introduces a \emph{multi-scale} effect: near the source, reflections come from a smaller region; farther away, reflections integrate over a larger area. Due to small apex angle, the physical contact region is considered as an ellipse in most cases. However, our beam tracing algorithms only returns a single hit-point ${\bf x}$ for each reflection. To model the overall reflection response from the area, we approximate it by a small Gaussian distribution around the sampled hit point ${\bf x}$. Concretely, we let ${\bf x'}\sim \mathcal{N}({\bf x}, \Sigma)$, where $\Sigma$ encodes the elliptical patch size and orientation. Intuitively, $\Sigma$ grows with the travel distance $l$ and depends on the reflection angle $\theta$ and half-apex angle $\varphi$. For the detailed calculation of $\Sigma$, please refer to Supp.

Next, we introduce our modeling of the reflection response.
Recall from Equation \ref{eq:path-freq} 
that $\mathrm{\rm Refl}({\bf x})$ is a frequency-dependent surface function. We discretize the frequency axis into $F$ key frequencies and learn reflection values at those key frequencies, linearly interpolating to produce a continuous frequency response. We thus seek:
\vspace{-0.05in}
\begin{equation}
    {\rm Refl}(\gamma({\bf x}, \Sigma); {\bf \Theta}_2) \in \mathbb{R}^F,\label{eq:refl-redef}
\vspace{-0.05in}
\end{equation}
where $\gamma(\mathbf{x}, \Sigma)$ is the integrated positional encoding (IPE) of surface location ${\bf x}$. We adopt the formulation of IPE from Mip-NeRF \cite{Barron2021MipNeRFAM} to handle the spread in the contact region, meaning that we replace the point-based Fourier positional encoding $\gamma({\mathbf{x}})$ with an ``averaged'' encoding:
\begin{align}
    \gamma({\bf x}, \Sigma) &= \mathbb{E}_{{\bf x'}\sim \mathcal{N}({\bf x}, \Sigma)} [\gamma({\mathbf{x}'})] \\
    &= \gamma(\mathbf{x})\circ \exp{\left(-\tfrac{1}{2}\mathrm{diag}(\Sigma_{\gamma})\right)}
\end{align}
Though not explicitly integrating over the elliptical region, this encoding incorporates the local variance of the beam contact, allowing reflection predictions at the same surface point to vary when the beam’s elliptical region differs.

\subsection{Multi-View Vision Feature Encoder}
\label{sec:vision}

We incorporate \emph{multi-view images} to guide reflection response prediction. Specifically, we learn a vision feature encoder $\mathrm{F}(\mathbf{x}, \Sigma;\mathbf{\Theta}_{2}')$ that captures the appearance information around the surface $\mathbf{x}$. The reflection response is then:
\vspace{-0.05in}
\begin{equation}
    \mathrm{Refl}\Big(
        \gamma({\bf x}, \Sigma), \mathrm{F}({{\bf x}, \Sigma}); \mathbf{\Theta}_2\Big) \in \mathbb{R}^F
\vspace{-0.05in}
\end{equation}

Assume we have a set of $N_s$ basis sample points on geometry $\mathcal{M}$, denoted by $\{\mathbf{z}_j\}_{j=1}^{N_s}$, and $N_c$ captured images, denoted by $\{I^{(i)}\}_{i=1}^{N_c}$. We construct $\mathrm{F}$ in three stages:
\begin{enumerate}[itemsep=0pt, parsep=0pt, topsep=0pt, partopsep=0pt]
    \item \emph{Per-view feature extraction.} This stage encodes each image into a feature map, then sample per-sample features from each image.
    \item \emph{Multi-view feature aggregation.} This stage aggregates multi-view features of each sample point $\mathbf{z}_j$ into a single feature vector $\mathbf{v}_j$.
    \item \emph{Sample-level neighborhood fusion.} This stage fuses the features of the $k$-nearest sample points around a query ${\bf x}$ into the unified vision feature vector $\mathrm{F}({\bf x}, \Sigma)$.
\end{enumerate}

\vspace{0.05in}
\noindent\textbf{Per-View Feature Extraction.} Let $I^{(i)}$ be the i-th image with known camera intrinsics $\pi$ and extrinsics $P^{(i)}$. We first extract a pixel-aligned feature map  $W^{(i)} = \mathcal{E}(I^{(i)})$ using a pre-trained vision encoder $\mathcal{E}$ (\eg, DINO-v2\cite{oquab2023dinov2}). Then, for each 3D sampled point $\mathbf{z}_j$, we project it into image $i$ with $P^{(i)}$ and $\pi$. We then bi-linearly sample from $W^{(i)}$ to obtain the vision feature $\mathbf{v}^{(i)}_j$:
\begin{equation}
    \mathbf{v}^{(i)}_j=\begin{cases}{W}^{(i)}\!\Bigl(\pi\left({P}^{(i)}\mathbf{z}_j\right)\Bigr)& \text{if $\mathbf{z}_j$ 
 is visible in ${I}^{(i)}$}, \\
        \mathbf{0} & \text{otherwise}.
    \end{cases}
\end{equation}

\vspace{0.05in}
\noindent\textbf{Multi-View Feature Aggregation.} Given the per-view feature ${\bf v}_{j}^{(i)}$, we next aggregate them across images to form a single vision feature ${\bf v}_j$ for each sample $\mathbf{z}_j$. We adopt cross-attention mechanism \cite{transformer}:
\vspace{-0.05in}
\begin{equation}
    \mathbf{v}_{j}^{\mathrm{raw}} = {\sum_{i=1}^{N_c}} {\rho}\Bigl({\phi(\gamma({\mathbf{z}_j}))^\top\psi(\mathbf{v}_{j}^{(i)}\oplus P^{(i)})} + m_{j}^{(i)}\Bigr) \cdot \alpha(\mathbf{v}_{j}^{(i)})
\vspace{-0.05in}
\end{equation}
where $\mathbf{v}_{j}^{\mathrm{raw}}$ is the intermediate aggregated feature. The functions $\psi$, $\phi$, and $\alpha$ are linear projections to produce keys, quires, and values, respectively. The mask $m_{j}^{(i)}$ is $0$ if $\mathbf{z}_j$ is visible in $I^{(i)}$ and $-\infty$ if otherwise. $\gamma(\cdot)$ is a Fourier positional encoding, and $\rho$ denotes the softmax function. Finally, a 2-layer feed-forward network (FFN) refines this raw feature:
\begin{equation}        
    \mathbf{v}_j =  \mathrm{FFN}(\mathbf{v}_{j}^{\mathrm{raw}}).
\end{equation}

\vspace{0.05in}
\noindent\textbf{Sample-Level Neighborhood Fusion.} With ${\bf v}_j$ computed for each sampled point $\mathbf{z}_j$, we then derive a vision feature for an arbitrary query point $\mathbf{x}$. Let
\begin{equation}
    N(\mathbf{x}) = \{\mathbf{z}^*_j\}_{j=1}^k
\end{equation}
be the set of $k$-nearest samples to $\mathbf{x}$, with corresponding vision features $\{\mathbf{v}^*_j\}_{j=1}^k$. We then fuse these neighborhood features using a point-transformer \cite{zhao2021point}:
\vspace{-0.05in}
\begin{equation}
    \mathrm{F}(\mathbf{x}, \Sigma) =\!\sum_{j=1}^k\rho\Bigl(g\bigl( \phi'(\gamma(\mathbf{x},\Sigma)) - \psi'(\mathbf{v}^*_j)+\delta_j\bigr)\Bigr) \bigl(\alpha'(\mathbf{v}_j^*)+\delta_j\bigr),
\vspace{-0.05in}
\end{equation}
where $g$ is a MLP producing the non-normalized attention coefficients from the key-query difference, and $\phi'$, $\psi'$, and $\alpha'$ are key, query, value projections. $\delta_j$ is a positional encoding produced by another MLP $\vartheta'$:
\vspace{-0.05in}
\begin{equation}
    \delta_j = \vartheta'(\mathbf{x} - \mathbf{x}_j^*).
\vspace{-0.05in}
\end{equation}

In this way, our two-level fusion across both views and sampled points yields a robust vision feature that captures local geometry, visibility, and appearance. These features then drive more accurate reflection response predictions in the subsequent acoustic modules.

\begin{table*}[ht]
    \centering
    \begin{tabular}{l c cccc cccc}
    \toprule
   \multirow{3}{*}{\textbf{Method}}&\multirow{3}{*}{\textbf{Scale}} &\multicolumn{4}{c}{\bfseries RAF-Empty}&\multicolumn{4}{c}{\bfseries RAF-Furnished }\\
   \cmidrule(lr){3-6} \cmidrule(lr){7-10}
       &  &Loudness & C50 & EDT & T60 &Loudness & C50 & EDT & T60  \\
        &   & (dB) $\downarrow$ & (dB) $\downarrow$ & (ms) $\downarrow$ & (\%) $\downarrow$  & (dB) $\downarrow$ & (dB) $\downarrow$ & (ms) $\downarrow$ & (\%) $\downarrow$\\
         \midrule
        NAF++ \cite{chen2024RAF, luo2022learning} & 1\%  & 6.05 &2.10&94.5&23.9 & 6.61 &2.10 & 74.9 & 23.0 \\
        INRAS++ \cite{chen2024RAF, su2022inras} & 1\%   & 3.69 &2.59&100.3&23.5 &2.96 &2.61&92.6 & 25.0\\
        AV-NeRF \cite{liang23avnerf} & 1\%   & 3.16 & 2.52 & 96.4 &  21.8  & 2.92 & 2.64  & 96.7 & 24.5  \\
        AVR \cite{avr} & 1\%  & \cellcolor{lemonchiffon}3.00 & 2.19& {87.3} & 24.1& 2.97 &  2.33 & \cellcolor{lemonchiffon}{72.3} & 17.9  \\
         \midrule
        Ours & 0.1\%  & 3.14 &\cellcolor{lemonchiffon}1.81&\cellcolor{lemonchiffon}86.6& \cellcolor{lemonchiffon}{16.9} & \cellcolor{lemonchiffon}2.45 &\cellcolor{lemonchiffon}1.98&80.1 &\cellcolor{lemonchiffon}{15.2}\\
        Ours & 1\%   & \cellcolor{grannysmithapple}2.50 &\cellcolor{grannysmithapple}{1.42}& \cellcolor{grannysmithapple}{56.2} &\cellcolor{grannysmithapple}{10.7}&\cellcolor{grannysmithapple}1.68 & \cellcolor{grannysmithapple}{1.29} & \cellcolor{grannysmithapple}{47.4} & \cellcolor{grannysmithapple}{9.61}   \\
        \bottomrule
    \end{tabular}
    \caption{
    Results on the Real Acoustic Field dataset~\cite{chen2024RAF} (0.32\,s, 16\,kHz). Cells highlighted in \textcolor{ForestGreen}{green} denote the best performance, and \textcolor{Dandelion}{yellow} indicates the second best. Note that our model trained on only 0.1\% of the data already achieves lower C50 and T60 errors than baseline methods, and significantly outperforms all baselines when using the same amount of training data.
    }
    \vspace{-0.1in}
    \label{tab:exp-raf}
\end{table*}

\subsection{Position-Dependent Residual Component}
\label{sec:residual}

Finally, we introduce the residual component $\mathrm{r}(t;\mathbf{\Theta}_3)$, which is responsible for modeling high-order reflections, diffuse reflections, diffraction, and late reverberations. We treat each point on the geometry $\mathbf{x}\in\mathcal{M}$ as a secondary sound source and model the residual IR as the integral of these secondary sources from the listener position $\mathbf{x}_b$.

We use a 4-layer MLP $\epsilon$, to predict the differential time-domain response $h(t)$ per solid angle $\omega$ at any location $\mathbf{x}$:
\begin{equation}
    \mathrm{h}(t) = \epsilon(\mathbf{x}, \omega, t, \mathbf{x}_a, \mathbf{p}_b; \mathbf{\Theta}_3).
\end{equation}
The residual IR is then calculated by the integral:
\vspace{-0.05in}
\begin{align}
    \mathrm{r}(t; \mathbf{\Theta}_3) &= \int_{\mathbb{S}^2}\mathcal{S}_{\tau}\{  \epsilon\}(\mathbf{x}'(\omega), -\omega, t; \mathbf{\Theta}_3)p_{\mathrm{u}}(\omega)d\omega \label{eq:method-full-res}\\
    &\approx \sum_{k=1}^{N_r}S_{\tau_k}\{\epsilon\}(\mathbf{x}'_k, -\omega_k, t;\mathbf{\Theta}_3).
\vspace{-0.2in}
\label{eq:method-mc-res}
\end{align}
In Equation \ref{eq:method-full-res}, $\mathbf{x}'(\omega)$ is the intersection of a ray in direction $\omega$ from $\mathbf{x}$ with room geometry $\mathcal{M}$, and $p_{\mathrm{u}}$ is the uniform distribution over $\mathbb{S}^2$. Equation \ref{eq:method-mc-res} approximates Equation \ref{eq:method-full-res} via Monte Carlo integration with $N_r$ sampled directions $\omega_k$ from distribution $p_{\mathrm{u}}$.

\section{Experiments}

\begin{table*}[ht]
   \small
    \centering
   \begin{tabular}{l ccc ccc ccc ccc}
    \toprule
   \multirow{3}{*}{\textbf{Method}}   &\multicolumn{3}{c}{\bfseries Classroom}&\multicolumn{3}{c}{\bfseries Complex Room} &\multicolumn{3}{c}{\bfseries Dampened Room}&\multicolumn{3}{c}{\bfseries  Hallway} \\
     \cmidrule(lr){2-4}
     \cmidrule(lr){5-7}
     \cmidrule(lr){8-10}
     \cmidrule(lr){11-13}
         & Loud & C50 & T60 & Loud & C50 & T60 & Loud & C50 & T60 & Loud & C50 & T60 \\
        & (dB) $\downarrow$ & (dB) $\downarrow$ & (\%) $\downarrow$ & (dB) $\downarrow$ & (dB) $\downarrow$ & (\%)$\downarrow$  & (dB) $\downarrow$ & (dB) $\downarrow$ & (\%) $\downarrow$ & (dB) $\downarrow$ & (dB) $\downarrow$ & (\%) $\downarrow$ \\
        \midrule
        NAF++ \cite{chen2024RAF,luo2022learning} & 8.27 & 1.62  & 134.0 & 4.43 & 2.25  & 44.8&  3.88 & 4.24 & 306.9 & 8.71 & 1.36 & 21.4 \\  
        INRAS++ \cite{chen2024RAF,su2022inras} &\cellcolor{lemonchiffon} 1.31 & 1.86 & 60.9 & \cellcolor{lemonchiffon}1.65 & 2.26  & 29.5 & 3.45 & 3.28 & 187.1& 1.55 &  1.87  & 7.4 \\
        AV-NeRF\cite{liang23avnerf} & 1.51 &  \cellcolor{lemonchiffon}1.43  & 50.0 & 2.01& \cellcolor{lemonchiffon}1.88  & 36.6& 2.40 & 3.05& 107.9 & \cellcolor{lemonchiffon}1.26 & \cellcolor{grannysmithapple} 1.03  &  9.5\\
        AVR \cite{avr} & 3.26 & 4.18  & 44.3 & 6.47 & 2.55 & 36.7 & 6.65 & 11.11 & 81.4 & 2.48 & 2.69   &  7.0\\
        Diff-RIR \cite{hearinganythinganywhere2024} & 2.24 & 2.42 & \cellcolor{lemonchiffon}39.7 & 1.75 & 2.23 & \cellcolor{lemonchiffon}18.5& \cellcolor{lemonchiffon}1.87 & \cellcolor{lemonchiffon}1.56  & \cellcolor{lemonchiffon}44.9 &  1.32 &   3.13  & \cellcolor{lemonchiffon}6.8 \\
        \midrule
        Ours& \cellcolor{grannysmithapple}0.99 & \cellcolor{grannysmithapple}1.02  & \cellcolor{grannysmithapple}24.3 & \cellcolor{grannysmithapple}0.98 & \cellcolor{grannysmithapple}1.44  & \cellcolor{grannysmithapple}10.8 & \cellcolor{grannysmithapple}1.11  & \cellcolor{grannysmithapple}1.45  & \cellcolor{grannysmithapple}31.9 & \cellcolor{grannysmithapple}0.85 & \cellcolor{lemonchiffon}1.15  & \cellcolor{grannysmithapple}6.3\\
        \bottomrule
    \end{tabular}
    \caption{
Results on the Hearing Anything Anywhere dataset~\cite{hearinganythinganywhere2024} (2.0\,s segments, 16\,kHz), trained on 12 listener locations. Our method significantly outperforms all baseline methods in these scenes, demonstrating its effectiveness in accurately reconstructing room acoustics in few-shot settings. See Supp. for EDT error results.
}
\vspace{0.01in}
    \label{tab:exp-diffrir1}
\end{table*}

\subsection{Experiment Settings}

\noindent\textbf{Datasets.} We evaluate our method on two real-world datasets: the \emph{Real Acoustic Field} (RAF)~\cite{chen2024RAF} dataset and the \emph{Hearing Anything Anywhere} (HAA)~\cite{hearinganythinganywhere2024} dataset, which are the only available real-world RIR datasets with accompanying visual capture. The RAF dataset contains densely measured monaural RIRs recorded in two office settings (\emph{Empty} and \emph{Furnished}) using tens of thousands of source–listener pairs. We use 0.32\,s RIR segments resampled at 16\,kHz, following prior work~\cite{chen2024RAF,avr}. Each room also has a 3D reconstruction and a dense set of images~\cite{VRNeRF}. 

The HAA dataset comprises four rooms with diverse structures and acoustic characteristics. Each room has manually crafted planar geometry. Following \cite{hearinganythinganywhere2024}, we train on 12 listener locations per room and test on half of the remaining unseen locations. For HAA, RIR segments of 2.0\,s are used, and they are resampled at 16\,kHz.

\vspace{0.05in}
\noindent\textbf{Evaluation Metrics.} Following \cite{chen2024RAF,avr,su2022inras}, we evaluate perception-related energy decay patterns using \emph{Clarity} (C50), \emph{Early Decay Time} (EDT), and \emph{Reverberation Time} (T60). To account for differences in overall RIR magnitude, we also adopt a loudness metric defined as:
\begin{equation}
    \text{Loudness Error} = \Bigl|10\log_{10}\bigl(\frac{E_{\mathrm{pred}}}{E_{\mathrm{gt}}}\bigr)\Bigr|,
\end{equation}
where \(E=\int_0^{\infty}h^2(t)\,\mathrm{d}t\) is the energy of the signal \(h(t)\).

\begin{figure}[t!]
    \centering
    \includegraphics[width=1\linewidth,draft=false]{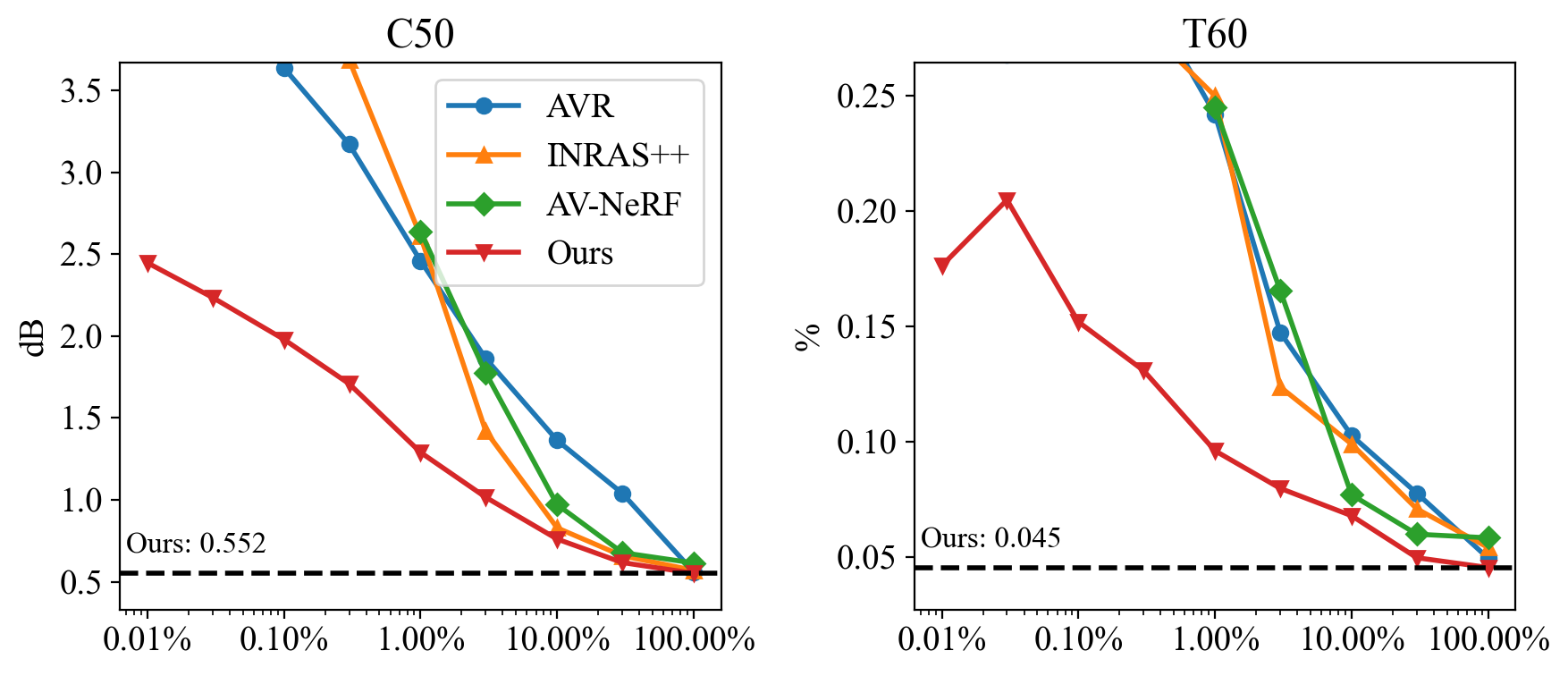}
    \caption{Performance comparison across training scales (from 0.01\% to 100\% of training data). Our model consistently outperforms baselines, particularly in few-shot scenarios (less than 3\% data). See Supp. for EDT and Loudness metrics.
    } 
    \label{fig:scale}
\end{figure}

\vspace{0.05in}
\noindent\textbf{Implementation Details.} For multi-view images, we manually select a subset covering the scene, using 13--30 images for most scenes in the two datasets and 65 images for the RAF Furnished room to assess performance saturation (see Supp. for ablation details). In the HAA dataset, where multi-view images are not provided, we randomly sample cameras from the Polycam reconstructions to render images at 512$\times$512 resolution. Note that Polycam is used only for visual rendering, while beam tracing is performed on the original coarse planar geometry. Additional architectural choices and training details are provided in Supp.

\vspace{0.05in}
\noindent\textbf{Baselines.} We compare with a series of state-of-the-art approaches. NAF++ and INRAS++~\cite{chen2024RAF} are improved versions of the implicit neural field methods NAF~\cite{luo2022learning} and INRAS~\cite{su2022inras} in 3D settings, respectively. AV-NeRF~\cite{liang23avnerf} is an audio-visual method that also uses depth and RGB to learn implicit acoustic fields. AVR~\cite{avr} is a physics-based method that uses NeRF-like volume rendering for RIR reconstruction. Diff-RIR~\cite{hearinganythinganywhere2024} is a differentiable rendering method that uses the image-source method for forward rendering; due to its extensive pre-computation time (over 500 hours for just two reflections in one scene on the RAF dataset), we only include it in the HAA comparisons.

\subsection{Quantitative Results}

\paragraph{Results on the RAF Dataset.}
To fully exploit the dense samples in RAF~\cite{chen2024RAF} and evaluate performance at various training scales, we split the original training set (80\% of all data) into 9 nested subsets ranging from 0.01\% to 100\% of the data. The smallest subset contains only 3 samples, while the largest includes approximately 30K samples, with each larger subset including all samples from the smaller ones. All models are evaluated on the original test set from \cite{chen2024RAF} to ensure comparability across scales. Table~\ref{tab:exp-raf} reports results on the 1\% dataset and on our model trained with 0.1\% of the data. Notably, our method trained on only 0.1\% of the data achieves comparable performance to state-of-the-art baselines trained on 10$\times$ more data. In addition, our method consistently outperforms all baselines when trained with the same amount of data, with improvements ranging from 16.6\% (Loudness Error in RAF Empty Scene) to 50.9\% (T60 Error in RAF Empty Scene). Figure~\ref{fig:scale} further demonstrates that our model outperforms existing baselines across all training scales, particularly in few-shot settings (below 3\% training data).

\begin{figure*}[ht!]
    \centering
    \includegraphics[width=0.99\linewidth,draft=false]{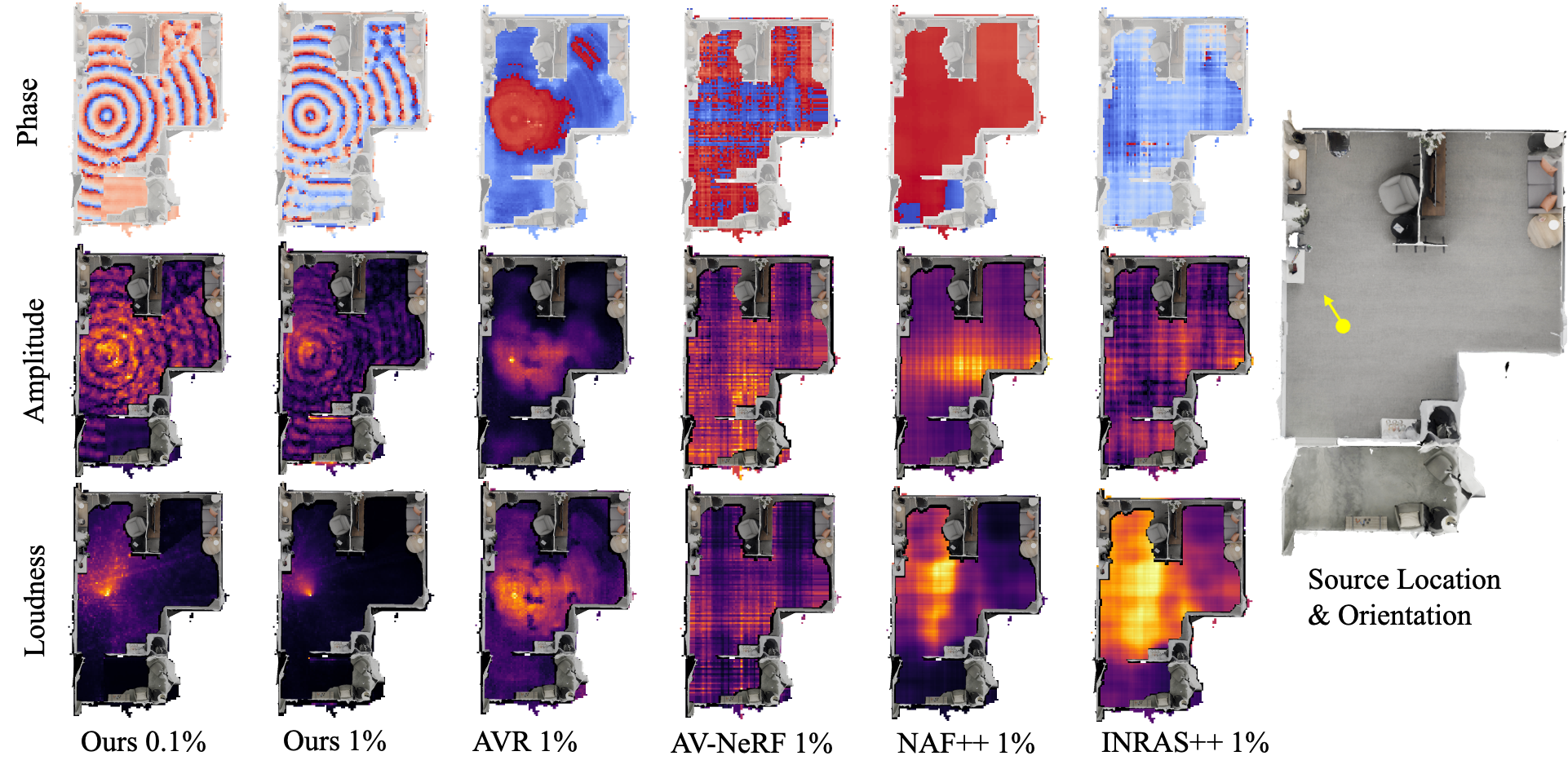}
        \vspace{-0.05in}
    \caption{
Signal spatial distribution visualization. \textbf{Top two rows:} Phase and amplitude maps at 0.6\,m wavelength. \textbf{Bottom row:} Loudness heatmap. Our model, trained on only 0.1\% of the data, accurately captures source directivity and localization, yielding plausible phase and amplitude distributions, while baseline methods fail to produce these patterns even with 10$\times$ training data.
    }
    \label{fig:wave-comparison}
\end{figure*}

\noindent\textbf{Results on the HAA Dataset.}
Table~\ref{tab:exp-diffrir1} shows the performance on the HAA dataset~\cite{hearinganythinganywhere2024}. Our method significantly outperforms all baseline methods in the four challenging real-world scenes, demonstrating its effectiveness in accurately reconstructing room acoustics. 
The only exception is the C50 metric in the \emph{Hallway} scene, where AV-NeRF exhibits particularly {strong} performance. This is likely due to AV-NeRF uses depth as an input, which is especially beneficial in this simple, constrained hallway geometry.

\noindent\textbf{Ablation Study.} We further conduct an ablation study where we ablate key components in our framework to evaluate their individual contributions: 1) \emph{Uni. Residual} denotes replacing the positional-dependent residual with a fixed learnable positional-independent vector; 2) \emph{w/o Residual} sets Residual to zero; 3) \emph{w/o Vision} sets vision features to zero; 4) \emph{Ray-tracing}
replaces beam tracing with specular ray tracing; and 5) \emph{w/o IPE} uses fixed Fourier features in place of the multi-scale positional encoding. The results are summarized in Table~\ref{tab:ablation}. We can see that each component is essential for accurately rendering RIRs at novel locations.

\begin{table}[t!]
    \centering
    \begin{tabular}{l cccc}
    \toprule
       Variant  &  C50 & EDT & T60 \\
    \midrule
        Ours (full) & 1.98 & 80.1 & 15.2  \\
        Uni. Residual & 2.11 & 106.4 & 13.9\\
        w/o Residual & 3.82 & 142.8 & 49.0 \\
        w/o Vision & 2.13 & 98.6 & 14.3\\
        Ray-tracing & 4.27 & 146.9&21.9 \\
        w/o IPE & 2.10 & 101.2 & 15.0\\
    \bottomrule
    \end{tabular}
    \vspace{0.01in}
    \caption{Ablation study results. See text for details.}
    \label{tab:ablation}
\end{table}

\subsection{Qualitative Results}

\begin{figure*}[t!]
    \centering
    \includegraphics[width=0.99\linewidth,draft=false]{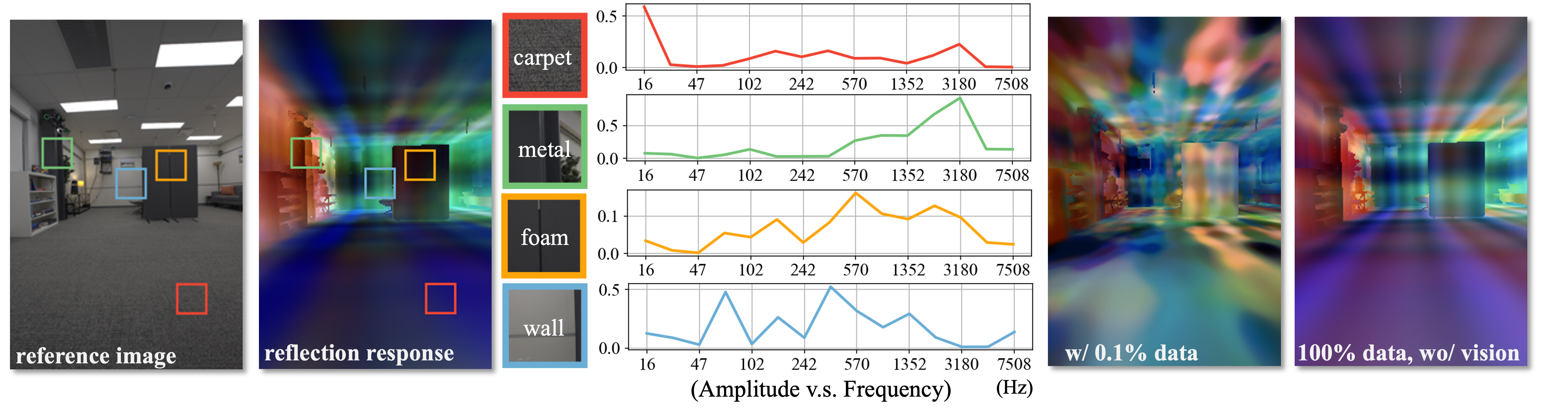}
    \vspace{-0.01in}
    \caption{
    Reflection response visualization. 
    The RGB color encodes frequency-dependent reflection response, with red indicating high-frequency and blue indicating low-frequency. Our method yields diverse, interpretable reflection patterns even with only 0.1\% training data. In the middle, we visualize the reflection response curves. The results align with real-world observations---\eg, carpet exhibits low high-frequency reflectivity, foam is generally absorptive, and metal reflects strongly at high frequencies.
    }
    \label{fig:reflection}
\end{figure*}

\noindent\textbf{Visualization of Signal Spatial Distribution.} Figure~\ref{fig:wave-comparison} illustrates the spatial distribution of the sound signal at an unseen source location and orientation
in the RAF Furnished Room. The top two rows display the phase and amplitude at a wavelength of 0.6\,m, while the bottom row shows the loudness heatmap. Our model generates physically plausible wave distributions, as evidenced by the periodic patterns in both phase and amplitude, and successfully captures the source directivity and reflection decay with as little as 0.1\% training data. In contrast, AVR, which is also a physics-based method, accurately models the source location but fails to capture the source directivity and produce physically consistent spatial distributions of phase and amplitude. Unsurprisingly, all learning-based methods (AV-NeRF, NAF++, INRAS++) struggle in this regard. While they can interpolate and predict RIRs by learning from, and sometimes overfitting to, the training data, they fail to produce a loudness heatmap with correct source localization or meaningful phase and amplitude distributions. For comparisons of predicted RIR signals, please refer to Supp.

\vspace{0.05in}

\noindent \textbf{Interpreting Learned Reflection Responses.} In Figure~\ref{fig:reflection}, we visualize the reflection response on the image surfaces by projecting RGB colors into the camera space. Red and blue indicate higher-frequency reflectivity and lower-frequency reflectivity, respectively.
Our method yields diverse and interpretable reflection responses, even when trained on just 0.1\% of the data. For example, the reflection responses for the carpet and metal areas align with their material properties---carpet effectively absorbs high frequencies, foam is generally absorptive, and metal is highly reflective in the high-frequency range. Moreover, an acoustic-only variant of our model (with vision features replaced by zeros) highlights the impact of incorporating visual cues, which significantly enhances the diversity and material relevance of the predicted reflection responses.

\section{Conclusion}

\vspace{-0.05in}
We presented \name, an audio-visual differentiable pipeline for synthesizing room impulse responses (RIRs). By combining beam tracing with visually-guided reflection modeling, our approach learns RIRs from sparse real-world measurements and outperforms state-of-the-art baselines while reducing data requirements. Our work opens new possibilities for immersive AR/VR applications. As future work, we plan to extend our framework to handle multi-scene scenarios for few-shot or zero-shot reflection response prediction. We also aim to explore implicit acoustic modeling from only raw audio data, leveraging much larger corpora for training more generalizable models.

{\small
\bibliographystyle{ieee_fullname}
\bibliography{egbib}
}

\clearpage

\supparxiv{
\setcounter{page}{1}
\maketitlesupplementary
}{}

\appendix

\renewcommand{\thesection}{A.\arabic{section}}
\setcounter{section}{0}

\section{Method Details}
\subsection{Acoustic Beam Tracing Algorithm}

\begin{figure}[ht]
    \centering
    \includegraphics[width=.8\linewidth, draft=false]{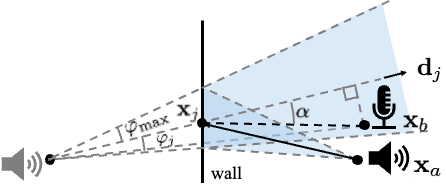}
    \caption{Acoustic beam tracing: in acoustic beam tracing the source and listener are considered as two point, the sound is propagate via a cone-shape beam in space. Acoustic beam tracing handles reflection the same as ray tracing does. The key difference is that acoustic beam tracing enumerate a reflection path if the listener is contatined in the beam volume but not necessarily being hitted by the sampled ray}
    \label{fig:hit-cond}
\end{figure}

Given the source location ${\bf x}_a$ and listener location ${\bf x}_b$, we adopt acoustic beam tracing~\cite{beam1,beam2,beam3,beam4} to sample specular beams in a source-to-listener manner. First we cast $N_d$ beams from the source, using a Fibonacci lattice \cite{Hardin2016ACO} to approximate uniform coverage of directions. A small apex angle $2\varphi_{\mathrm{max}}$ is selected to ensure the cone-shape beams remain disjoint. Next, each beam's center ray intersects with room geometry to find reflection points (e.g. via Open3D \cite{open3d}), and after each reflection, we check if the reflected beam can hit the listener. To determine whether a reflected beam at $j$-th reflection point $\mathbf{x}_j$ (with out-going direction $\mathbf{d}_j$) reaches the listener before hitting another surface, we check if the listener is within the reflected cone (as show in Figure \ref{fig:hit-cond}). Denote $l_j$ as the distance traveled by reaching $\mathbf{x}_j$, and $\alpha_j$ as the angle between $\mathbf{d}_j$ and the line from $\mathbf{x}_j$ to $\mathbf{x}_b$ and $\varphi_j$ as the sampled half-apex angle:
\begin{equation}
    \varphi_j = \arctan\left(\frac{\|{\bf x}_b - {\bf x}_j\|\sin\alpha}{\|{\bf x}_b - {\bf x}_j\|\cos\alpha + l_j}\right).
\end{equation}
The listener is considered ``hit'' if $\alpha$ is acute, $\varphi_j<\varphi_{\max}$, and $\mathbf{x}_j$ is visible by $\mathbf{x}_b$. In addition, the time-of-arrival is by:
\begin{equation}
    \mathrm{toa}_j = \frac{\|{\bf x}_b - {\bf x}_j\|\sin\alpha}{v_{\mathrm{sound}}\cdot\sin\varphi_j}.
\end{equation}
Algorithm \ref{algo:dbt} summarizes our beam-tracing procedure.

\begin{algorithm}[t]
\SetAlgoNlRelativeSize{-1}
\SetAlgoSkip{5pt}
    \DontPrintSemicolon
    \caption{Acoustic Beam Tracing}
    \label{algo:dbt}
    \KwIn{Source ${\bf x}_a$, Listener ${\bf x}_b$, Geometry $\mathcal{M}$}
    \KwOut{Specular paths $\{\tilde{\bf x}_k\}_{k=1}^{N}$}
    
    \For{$i=1$ to $N_d$}{
        ${\bf x}_{i,0} \!\leftarrow\! {\bf x}_a$; 
        $l_{i,0} \!\leftarrow\! 0$;\;
        ${\bf d}_{i,0} \!\leftarrow \texttt{SampleFib}(N_d,i)$\;
    }
    \texttt{ANS} $\!\leftarrow\! \{\}$\;
    \If{\texttt{IsVisible}(${\bf x}_a$, ${\bf x}_b$)}{
        \texttt{ANS.add}($\varnothing$) \quad // direct path 
    }
    
    \For{$j=1$ to $\mathrm{MAX}_{\mathrm{depth}}$}{
        \For{$i=1$ to $N_d$}{
            $[{\bf x}_{i,j}, {\bf z}] 
              = \texttt{HitPoint}(\mathcal{M}, {\bf x}_{i,j-1}, {\bf d}_{i,j-1})$\;
            ${\bf d}_{i,j} 
              = {\bf d}_{i,j-1} - 2\,({\bf z}^\top {\bf d}_{i,j-1})\, {\bf z}$\;
            $l_{i,j} 
              = l_{i,j-1} 
                + \|{\bf x}_{i,j} - {\bf x}_{i,j-1}\|$\;
            
            \If{\texttt{BeamHit}(${\bf x}_b$, ${\bf x}_{i,j}$, ${\bf d}_{i,j}$, $l_{i,j}$)}{
                \texttt{ANS.add}$\bigl([\,{\bf x}_{i,1}, {\bf x}_{i,2},\ldots, {\bf x}_{i,j}]\bigr)$\;
            }
        }
    }
    \Return \texttt{ANS}\;
\end{algorithm}

\begin{figure}[ht]
    \centering
    \includegraphics[draft=false,width=.9\linewidth]{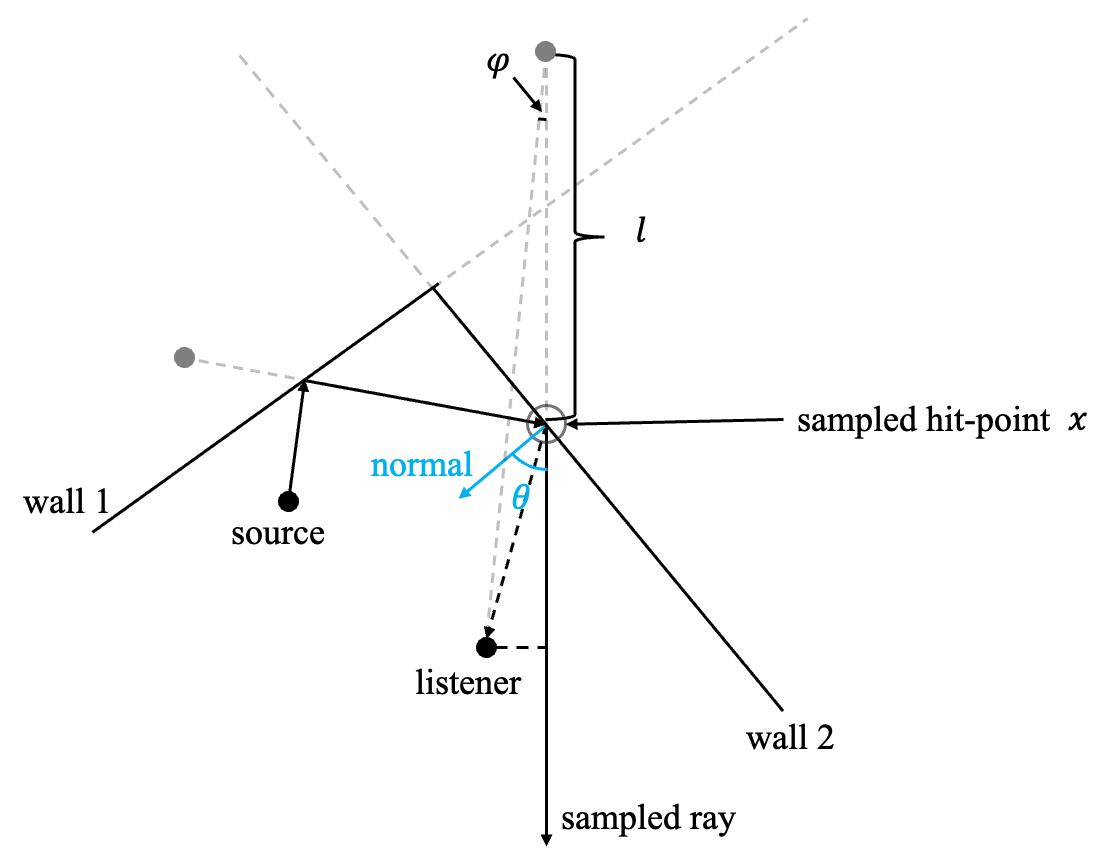}
    \caption{
    Local covariance derivation: as the traveling space $l$ increases, the region of the contact area expand linearly in terms of radius. In addition, since the half-apex angle is assumed to be small, the contact region is considered an ellipse, which motivates use model the region information with a gaussian distribution. 
    }
    \label{fig:multiscalerir}
\end{figure}

\subsection{Local Variance Derivation}

\begin{table*}[ht]
    \centering
    \begin{tabular}{lccccccccc}
    \toprule
    Room & Floor Area & $N_{\text{bounce}}$ & $N_{\text{basis}}$ & $L_{\text{RIR}}$ & $T_{\text{inference}}$  & $T_{\text{tracing}}$ & $T_{\text{res}}$ & $T_{\text{train}}$ & $N_{\text{params}}$\\
    \midrule 
    HAA-Classroom & $\sim$56$\text{m}^2$ & 6 & 2.3K & 2.00s & 66.0ms & 30.4ms & 11.2ms & 0.59h & 26.4M \\
    HAA-Complex &$\sim$106$\text{m}^2$ & 6 & 5.1K & 2.00s & 69.6ms & 32.5ms & 11.5ms & 1.04h & 26.4M \\
    HAA-Dampened &$\sim$25$\text{m}^2$ & 2 & 1.1K & 2.00s & 26.8ms & 9.58ms & 11.5ms & 0.36h & 26.4M \\
    HAA-Hallway &$\sim$28$\text{m}^2$ & 6 & 1.7K & 2.00s & 67.6ms &29.0ms & 11.2ms & 1.86h & 26.4M\\
    RAF-Furnished & $\sim$44$\text{m}^2$ & 4 & 5.9K & 0.32s & 51.3ms & 35.5ms & 4.12ms & 6.63h & 19.5M\\
    RAF-Empty & $\sim$44$\text{m}^2$ & 4 & 4.9K & 0.32s &  54.4ms & 35.1ms & 4.07ms & 10.8h & 19.5M \\
    \bottomrule
    \end{tabular}
    \caption{Detailed computational breakdown across HAA and RAF scenes. Here, $N_{\text{bounce}}$ denotes the number of reflections simulated per beam, $N_{\text{basis}}$ is the basis points sampled as descributed in \S\ref{ssec:basis}, $L_{\text{RIR}}$ is the RIR duration, and $N_{\text{params}}$ is the total number of trainable parameters in our model. All RIR are sampled at 16kHz.
    Training times for RAF scenes are reported with 1\% training data.}
    \label{tab:inftime-only}
\end{table*}

As shown in Figure \ref{fig:multiscalerir}, consider a beam traveling distance $l$ before hitting the surface at $\mathbf{x}$, with half-apex angle $\varphi$ and local surface normal $\mathbf{z}$. Let $\theta$ be the angle between the reflected direction $\mathbf{d}$ and $\mathbf{z}$. In a local coordinate system whose axes are $\{\mathbf{t}_1, \mathbf{t}_2, \mathbf{z}\}$, where we requires $\mathbf{t}_1$ aligns with the projection of $\mathbf{d}$ in the tangent surface, the beam’s cross-section at distance $l$ is approximately an ellipse with semi-major and semi-minor axes proportional to \(l \sin \varphi\), modulated by \(\theta\). A simple way to encode this elliptical patch is to use a diagonal covariance at local coordinate
\begin{equation}
  \Sigma_{\mathrm{local}}
  =
  \mathrm{diag}\!\Bigl(
    \sigma_{1}^2,\;\sigma_{2}^2,\;0
  \Bigr),
\end{equation}
where $\sigma_{1}^2$ and $\sigma_{2}^2$ grow with \(l\sin\varphi\), adjusted by $\cos\theta$. In the case when $\varphi$ is small:
\[
    \sigma_{1}^2 \,\approx\, \bigl(l\sin\varphi\bigr)^2/\cos^2\!\theta,
    \quad
    \sigma_{2}^2 \,\approx\, \bigl(l\sin\varphi\bigr)^2/\cos\!\theta.
\]
These terms capture how the beam’s ellipse “stretches” along \(\mathbf{t}_1\) and \(\mathbf{t}_2\). In world coordinates, the final covariance \(\Sigma\) is simply
\[
  \Sigma \;=\;
  Q\,\Sigma_{\mathrm{local}}\,Q^\top,
\]
where \(Q=[\,\mathbf{t}_1\,\mathbf{t}_2\,\mathbf{z}\,]\) rotates from local axes to world axes.

\begin{figure}
    \centering
    \includegraphics[width=1\linewidth,draft=false]{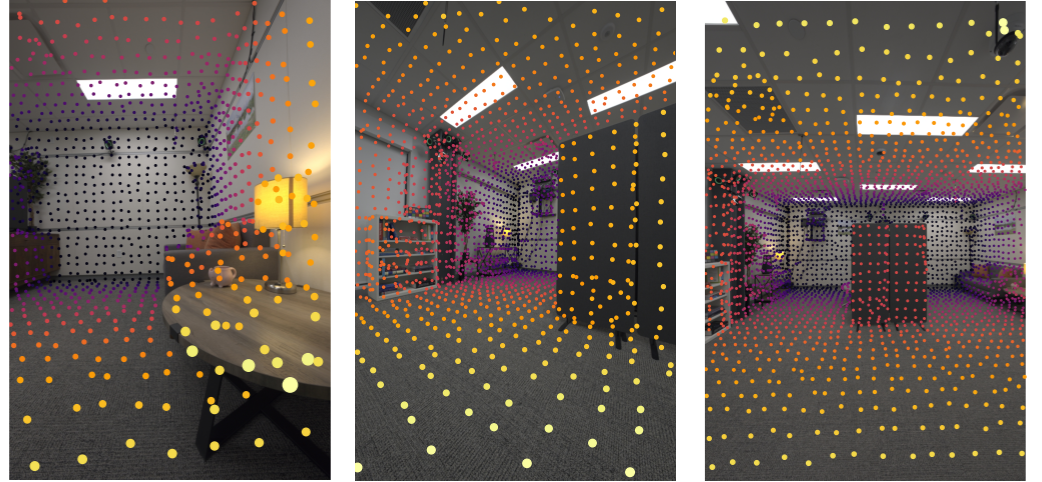}
    \caption{Visualization of surface basis samples for extracting multi-view images features.}
    \label{fig:basis}
\end{figure}

\subsection{Basis Points Sampling}
\label{ssec:basis}

we sample the basis point in two steps, first we densely sample 100,000 points on the room geometry, then, we downsample them with voxel size 0.2m and use the median point (closest to mean point) as the basis samples for vision features, as shown in Figure~\ref{fig:basis}, in this way, we ensures the distances between samples are stable.

\subsection{Hyperparameters}

Following \cite{hearinganythinganywhere2024}, we use a spherical Gaussian weighting function with a sharpness parameter of 8 for source directional response. We decode the image feature using a 4-layer MLP and sample frequencies from 12 to 7800\,Hz with 16 logarithmically spaced samples, linearly interpolating the frequency response.

\subsection{Optimization}
We optimize the network using the Adam optimizer with a fixed learning rate of \(5\times10^{-4}\) (and \(1\times10^{-4}\) for the residual component). Our loss function is defined as:
\begin{equation}
    \mathcal{L} = \mathcal{L}_{\mathrm{MAG}} + \lambda_{\mathrm{pink}} \mathcal{L}_{\mathrm{pink}} + \lambda_{\mathrm{decay}}\mathcal{L}_{\mathrm{decay}},
\end{equation}
where \(\mathcal{L}_{\mathrm{MAG}}\) is a multi-scale log L1 loss, \(\mathcal{L}_{\mathrm{pink}}\) is the pink noise supervision loss, and $\mathcal{L}_{\mathrm{decay}}$ is the decay loss proposed by~\cite{Liang2023NeuralAC,fewavlearningenvacoustics}. We adopt a progressive training strategy, starting with a reflection order \(N=1\) and increasing by 1 every 100 epochs until \(N=6\). During training, we sample 16,384 points from Fibonacci lattices for beam tracing, reducing this to 8,192 points per RIR during inference. Training is performed with a batch size of 128.

\subsection{Computational Cost}

\paragraph{Training/Inference Time and Model Size.}
We measure our model's size and training/inference time against baseline methods, as shown in Table~\ref{tab:inftime}. On the HAA Classroom dataset base setting, inference on a 2.0s, 16kHz RIR using a single RTXA6000 takes 66ms. Our approach achieves the fastest inference among existing physics-based methods (\ie, DiffRIR and AVR). 

\begin{table}[ht]
    \centering
    \begin{tabular}{lccc}
    \toprule
        Method & $T_{\text{inference}}$ & $T_{\text{training}}$ & Size \\
        \midrule
        DiffRIR &  376 ms  & 3.55 h & 34.4 K \\
        AVR & 100 ms & 0.17 h & 45.7 M \\
        Ours  & 66 ms & 0.59 h & 26.4 M\\
        \bottomrule
    \end{tabular}
    \caption{Training time, inference time, and model size comparison on HAA Classroom (2.0s IR, 16kHz, same RTX A6000)}
    \label{tab:inftime}
    \vspace{-0.15in}
\end{table}

\paragraph{Scene-Level Breakdown.}
Table~\ref{tab:inftime-only} decomposes the computation cost across additional HAA and RAF scenes. For each scene, we report the total inference time $T_{\text{inference}}$, which includes beam-tracing ($T_{\text{tracing}}$), residual rendering ($T_{\text{res}}$), and early-stage RIR rendering\footnote{We do not include it in the comparison, as isolating it would require caching the full beam trace for every source-listener pair, which is prohibitively memory intensive.}. 
Inference time remains below 70ms for all HAA rooms and below 55ms for the RAF dataset, demonstrating that our model naturally extends with scene complexity. As an additional sanity check, we also test our framework on a much larger scene---Gibson Hennepin~\cite{xiazamirhe2018gibsonenv} ($\sim$600$\text{m}^2$; 69k points; 6 bounces), our model costs 108ms per 8s, 16kHz RIR render.

\begin{figure*}[ht]
    \centering
    \includegraphics[width=1\linewidth,draft=false]{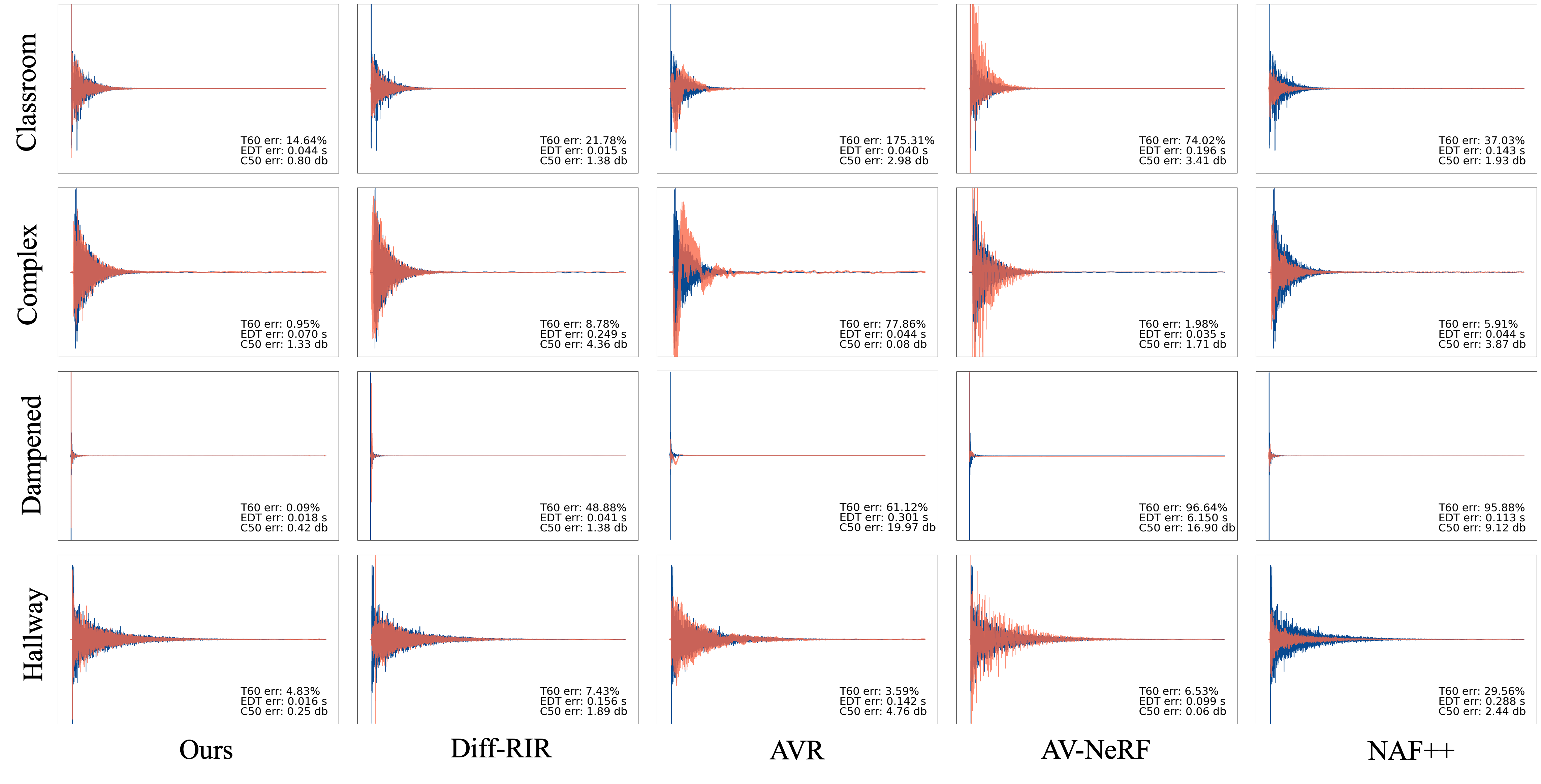}
    \caption{Wave visualization on the Hearing Anything Anywhere dataset~\cite{hearinganythinganywhere2024}. All models are trained on 12 data points. Our model significantly outperforms all baselines in preserving the wave structure—producing the most faithful wave front with accurate peak locations and magnitudes. Note that quantitative metrics do not always capture these perceptual details; some methods may have low error values despite producing distorted wave patterns.
    }
    \label{fig:haa-wave}
\end{figure*}
\begin{figure*}[ht]
    \centering
    \includegraphics[width=1\linewidth,draft=false]{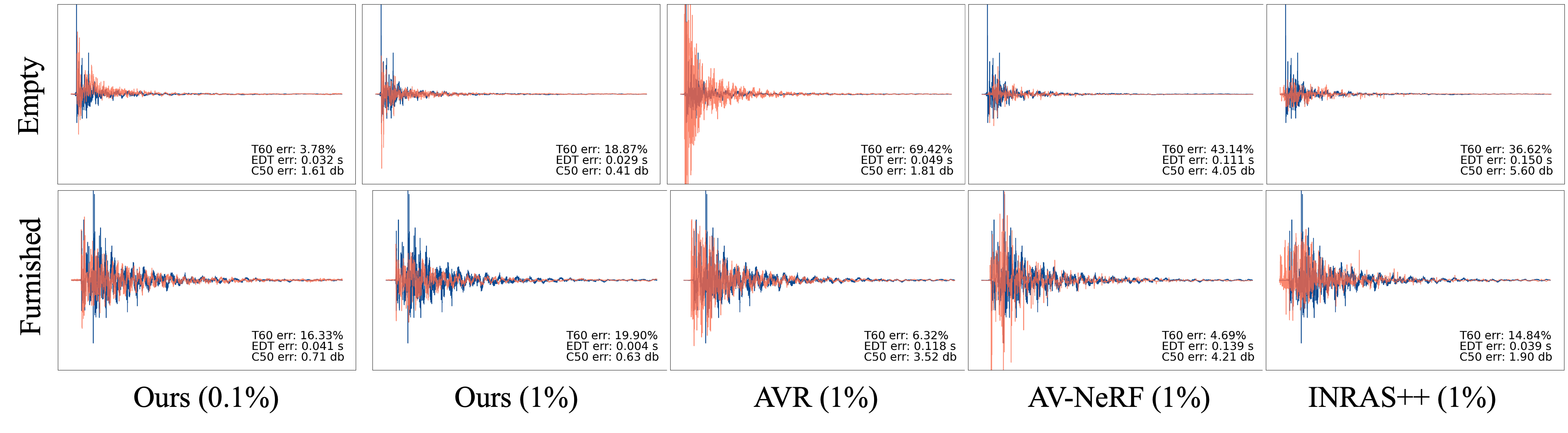}
    \caption{
    Wave visualization on the Real Acoustic Field dataset~\cite{chen2024RAF}. We show results from three baseline models trained on 1\% of the data alongside our model trained on 1\% and 0.1\% of the data. Our model exhibits better peak alignment and magnitude than baseline methods—even when trained on only 0.1\% of the data—and significantly outperforms all baselines when using the same amount of training data.
    }
    \label{fig:raf-wave}
\end{figure*}

\begin{table*}[ht!]
    \centering
   \begin{tabular}{l cccc cccc}
    \toprule
   \multirow{3}{*}{\textbf{Method}}   &\multicolumn{4}{c}{\bfseries Classroom}&\multicolumn{4}{c}{\bfseries Complex Room} \\
     \cmidrule(lr){2-5}
     \cmidrule(lr){6-9}
    
         &Loudness & C50 & EDT & T60 &Loudness & C50 & EDT & T60   \\
        & (dB) $\downarrow$ & (dB) $\downarrow$ & (ms) $\downarrow$ & (\%) $\downarrow$ & (dB) $\downarrow$ & (dB) $\downarrow$ & (ms) $\downarrow$ & (\%) $\downarrow$ \\
         \midrule
        NAF++ \cite{luo2022learning} & 8.27 & 1.62 & 162.3 & 134.0 & 4.43 & 2.25 & 203.5 & 44.8 \\  
        INRAS++ \cite{su2022inras} &\cellcolor{lemonchiffon} 1.31 & 1.86 & 110.0 & 60.9 & \cellcolor{lemonchiffon}1.65 & 2.26 & 150.7 & 29.5 \\
        AV-NeRF\cite{liang23avnerf} & 1.51 &  \cellcolor{lemonchiffon}1.43 &\cellcolor{lemonchiffon} 77.8 & 50.0 & 2.01& \cellcolor{lemonchiffon}1.88 & \cellcolor{lemonchiffon}107.9 & 36.6\\
        AVR \cite{avr} & 3.26 & 4.18 & 135.6 & 44.3 & 6.47 & 2.55 & 138.3 & 36.7 \\
        Diff-RIR \cite{hearinganythinganywhere2024} & 2.24 & 2.42 & 139.7 & \cellcolor{lemonchiffon}39.7 & 1.75 & 2.23 & 129.5 & \cellcolor{lemonchiffon}18.5 \\
        \midrule
        Ours& \cellcolor{grannysmithapple}0.99 & \cellcolor{grannysmithapple}1.02 &\cellcolor{grannysmithapple} 55.5 & \cellcolor{grannysmithapple}24.3 & \cellcolor{grannysmithapple}0.98 & \cellcolor{grannysmithapple}1.44 & \cellcolor{grannysmithapple}86.5 & \cellcolor{grannysmithapple}10.8 \\
        \bottomrule
    \end{tabular}
\vspace{0.05in}
    \begin{tabular}{l cccc cccc}
    \toprule
    \multirow{3}{*}{\textbf{Method}} &\multicolumn{4}{c}{\bfseries Dampened Room}&\multicolumn{4}{c}{\bfseries  Hallway} \\
     \cmidrule(lr){2-5}
     \cmidrule(lr){6-9}
    
         &Loudness & C50 & EDT & T60 &Loudness & C50 & EDT & T60   \\
         & (dB) $\downarrow$ & (dB) $\downarrow$ & (ms) $\downarrow$ & (\%) $\downarrow$ & (dB) $\downarrow$ & (dB) $\downarrow$ & (ms) $\downarrow$ & (\%) $\downarrow$ \\
         \midrule
        NAF++ \cite{luo2022learning}  &  3.88 & 4.24 & 360.0 & 306.9 & 8.71 & 1.36 & 148.3 & 21.4 \\
        INRAS++ \cite{su2022inras} & 3.45 & 3.28 & 187.1& 382.9 & 1.55 &  1.87 & 157.9 & 7.4\\
        AV-NeRF \cite{liang23avnerf} & 2.40 & 3.05 &  242.1 & 107.9 & \cellcolor{lemonchiffon}1.26 & \cellcolor{grannysmithapple} 1.03 &  \cellcolor{grannysmithapple} 89.9 &  9.5  \\
        AVR \cite{avr} & 6.65 & 11.11 & 305.3 & 81.4 & 2.48 & 2.69 & 195.4  &  7.0\\
        Diff-RIR \cite{hearinganythinganywhere2024} & \cellcolor{lemonchiffon}1.87 & \cellcolor{lemonchiffon}1.56 & \cellcolor{lemonchiffon}153.0 & \cellcolor{lemonchiffon}44.9 &  1.32 &   3.13 & 188.1 & \cellcolor{lemonchiffon}6.8\\
        \midrule
        Ours& \cellcolor{grannysmithapple}1.11  & \cellcolor{grannysmithapple}1.45 & \cellcolor{grannysmithapple}139.0 & \cellcolor{grannysmithapple}31.9 & \cellcolor{grannysmithapple}0.85 & \cellcolor{lemonchiffon}1.15 & \cellcolor{lemonchiffon}96.5 & \cellcolor{grannysmithapple}6.3 \\
        \bottomrule
    \end{tabular}
    \caption{{Result on the HAA~\cite{hearinganythinganywhere2024} dataset}, 2.0s, 16K sample rate}
    \label{tab:exp-diffrir1-supp}
\end{table*}

\section{Additional Results}

\subsection{Waveform Comparison}

Figure~\ref{fig:haa-wave} shows wave visualizations on the Hearing Anything Anywhere dataset. All models were trained on only 12 data points. Our model significantly outperforms the baselines in preserving the wave structure, producing a wave front that closely matches the ground truth in terms of peak locations and magnitudes. Note that quantitative metrics do not always capture these perceptual differences; some methods may achieve low error values despite generating distorted wave patterns. This comparison highlights the superior capability of our approach in modeling acoustic dynamics in few-shot settings.

Figure~\ref{fig:raf-wave} presents wave visualizations on the Real Acoustic Field dataset. Here, we compare three baseline models trained on 1\% of the data with our model trained on both 1\% and 0.1\% of the data. Our results demonstrate that, in terms of wave structure, our model achieves better peak alignment and peak magnitude than the baselines—even when our model is trained on only 0.1\% of the data. When trained on 1\% of the data, our method further outperforms the baselines.

\subsection{Multi-scale Performance Comparison}

\begin{figure}
    \centering
    \includegraphics[width=1\linewidth,draft=false]{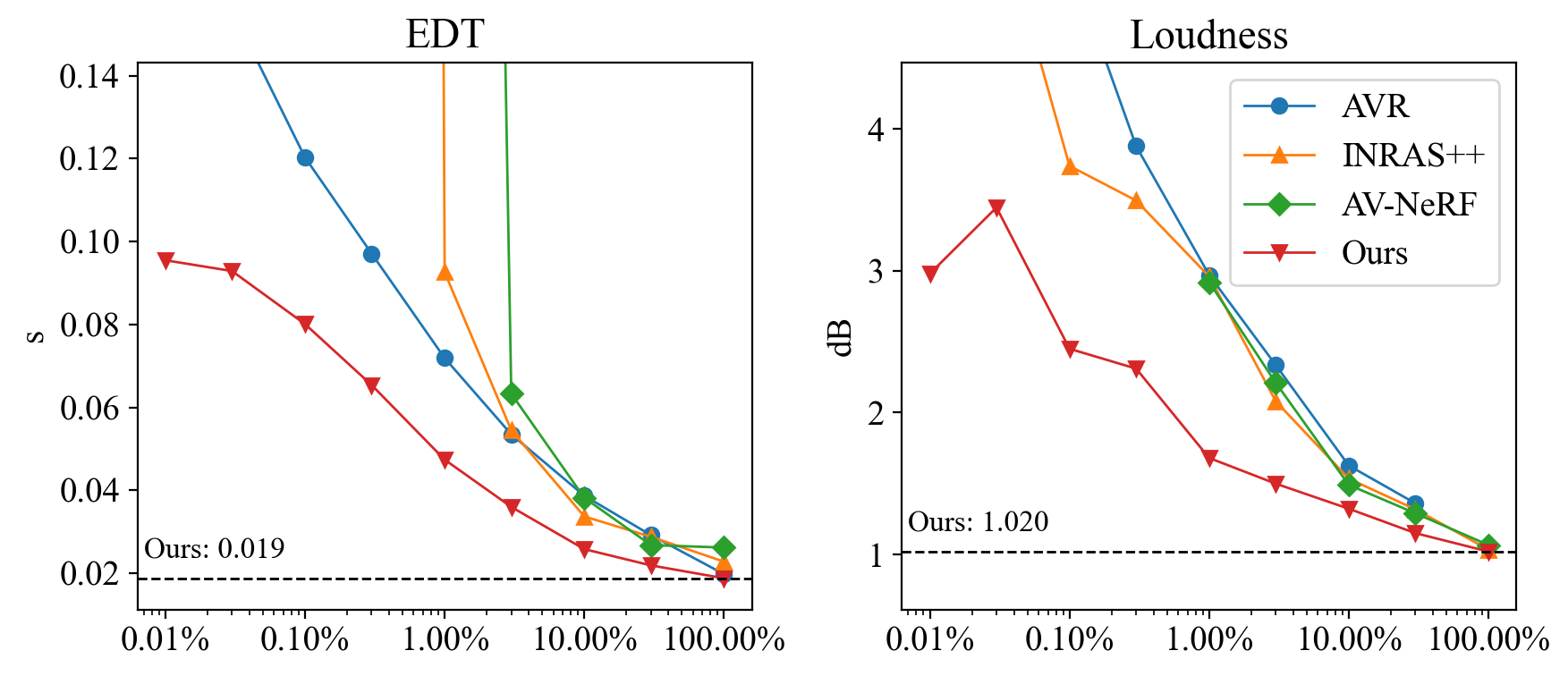}
    \caption{
    Performance comparison across training scales (from 0.01\% to 100\% of training data). In addition to the metrics reported in the main paper, our model consistently outperforms the baselines in terms of both EDT and Loudness.
    }
    \label{fig:scale-0}
\end{figure}

\paragraph{Loudness and EDT Errors.}
Figure~\ref{fig:scale-0} extend the multi-scale performance comparison in main paper by evluating on two more metrics, i.e., Loudness and EDT. The result shows that our model performs consistantly better than baselines in all training data scale, which is aligned with our observation in the main paper.

\paragraph{Initial Drop in T60 and Loudness Errors.} We discovered one of 9 RIRs in the 0.0003\% subset was invalid due to speaker failure, resulting in an almost silent recording.Excluding it, the T60 and Loudness errorss (15.7\% and 2.74dB, respectively) restore the expected monotonic decrease with larger dataset size. Only $0.27\%$ of RAF data were similarly affected; all other conclusions remain valid.

\subsection{Full Metrics on the HAA Dataset}

Table~\ref{tab:exp-diffrir1-supp} present the complete evaluation metrics on the HAA dataset, including Loudness, C50, EDT, and T60 across four scenes. Our results show that our method outperforms state-of-the-art baselines across almost all metrics, confirming the trends observed in the main paper. The only exception is the C50 metric and EDT metric in the \emph{Hallway} scene, where AV-NeRF performs particularly well, likely due to its effective use of depth information in this constrained geometry. These comprehensive results validate the robustness and effectiveness of our model in diverse real-world acoustic environments.

\begin{table}[ht]
    \centering
    \begin{tabular}{l ccc}
    \toprule
       Variant  &  C50 & EDT & T60 \\
    \midrule
        65 views & 1.98 & 80.1 & 15.2  \\
        20 views & 2.01 & 80.9 & 15.7 \\
        10 views & 2.13 & 97.9 & 15.2 \\
        5 views  & 2.12 & 97.2 & 15.3 \\
        ResNet18 & 1.96 & 89.4 & 15.3 \\
    \bottomrule
    \end{tabular}
    \caption{Ablation study on vision features. ``65 views'' denotes using 65 images for training; ``20 views'', ``10 views'', and ``5 views'' denote reduced image sets. ``ResNet18'' indicates replacing the DINO-V2 encoder with ResNet18.}
    \label{tab:view}
\end{table}

\subsection{Ablations on Vision Features}

We investigate the impact of vision features by varying two aspects: the number of multi-view images used for training and the choice of the pretrained encoder. Both experiments are conducted on the RAF Furnished scene using only 0.1\% of the training data.

Table~\ref{tab:view} shows our vision feature saturation experiment, we initially use 65 images to cover the entire scene, then reduce the number to 20, 10, and 5 views (see top four rows of Table~\ref{tab:view}). Reducing from 65 to 20 views incurs less than a 1\% drop in C50 and EDT, but further reduction from 20 to 10 views causes a marked performance decline, indicating that adequate view redundancy is essential for effective visual guidance. Performance remains stable when further reduced from 10 to 5 views, suggesting that with only 10 views the model nearly abandons visual feature learning and relies primarily on acoustic cues.

We also replace the DINO-v2~\cite{oquab2023dinov2} encoder with ResNet18~\cite{resnet}, which results in a noticeable drop in EDT, demonstrating that DINO-V2 is better suited for our model. Notably, all vision ablations have minimal impact on T60, indicating that vision features primarily contribute to modeling early reflection rather than late reverberation.

\subsection{Detailed Analysis on Model Components}

Table~\ref{tab:ablation-supp} summarizes the impact of ablating individual components of our model. 

\vspace{0.01in}
\noindent\textbf{Differentiable Renderer.} Replacing the learned residual field with a position-independent on (\emph{Uni. Residual}) increases EDT error by more than 30\%, and removing the residual entirely (\emph{w/o Residual}) raises C50, EDT, and T60 errors by over 70\%. Substituting our beam-tracing method with conventional ray tracing (\emph{Ray-Tracing}) worsens all three metrics by more than 40\%. Using plain Fourier features instead of integrated positional encoding (\emph{w/o IPE}) raises EDT by 26.3\%. 

\begin{table}[ht]
    \centering
    \begin{tabular}{l cccc}
    \toprule
       Variant  &  C50 & EDT & T60 \\
    \midrule
        Ours (full) & 1.98 & 80.1 & 15.2  \\
        Uni. Residual & 2.11 & 106.4 & 13.9\\
        w/o Residual & 3.82 & 142.8 & 49.0 \\
        w/o Vision & 2.13 & 98.6 & 14.3\\
        Ray-Tracing & 4.27 & 146.9&21.9 \\
        w/o IPE & 2.10 & 101.2 & 15.0\\
    \bottomrule
    \end{tabular}
    \caption{Ablation study results. See text for details.}
    \vspace{-0.05in}
    \label{tab:ablation-supp}
\end{table}

\vspace{0.01in}
\noindent\textbf{Vision Encoder. } Replacing vision encoder by zero vectors (\emph{w/o Vision}) degrades EDT error by roughly 23\%, confirming the importance of importance of visual information for accurate acoustic estimation.

These results confirm that each design choice contributes substantially to the overall performance.

\subsection{Failure Cases}
\begin{figure}[ht]
    \centering
    \includegraphics[width=1\linewidth,draft=false]{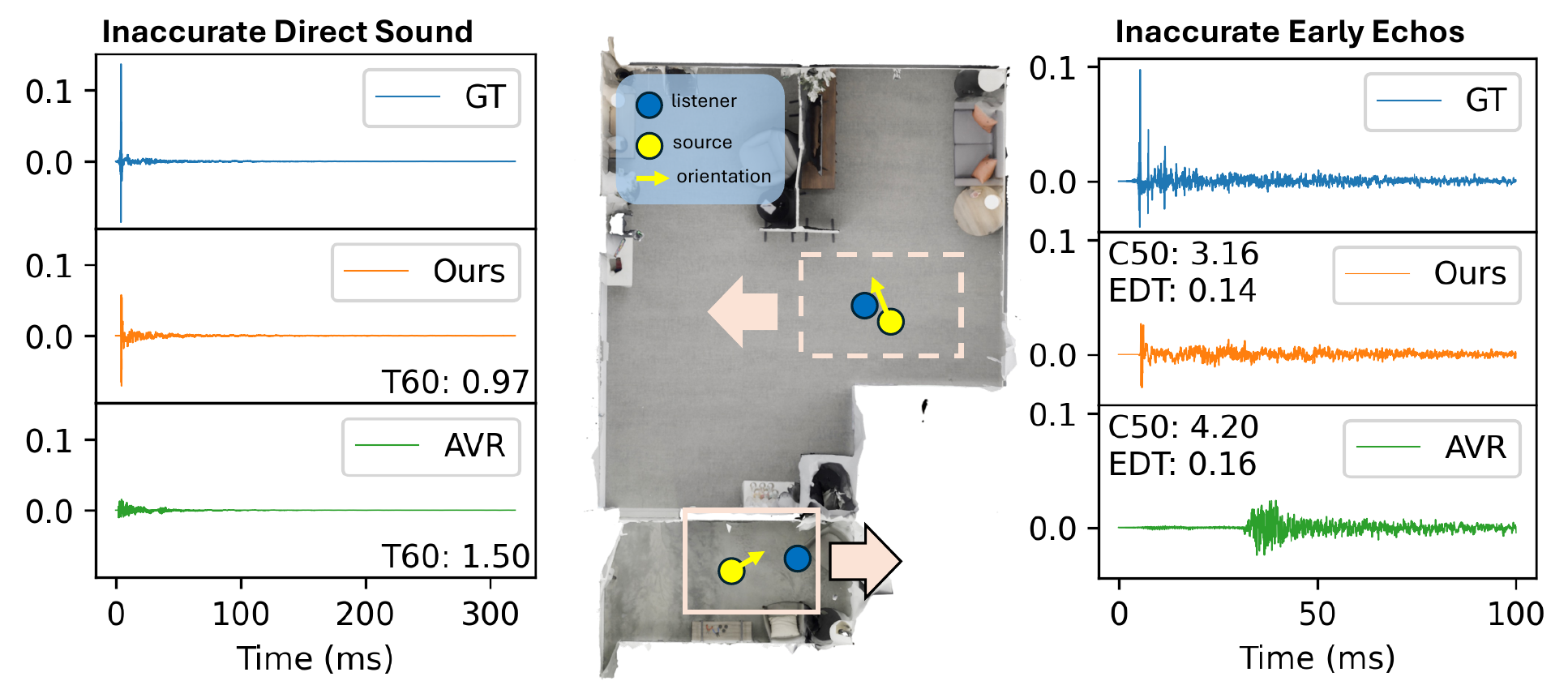}
    \caption{We visualize two failure cases in our model on RAF-Furnished Room with 1\% training data.}
    \label{fig:supp-failurecase1}
\end{figure}

In Figure~\ref{fig:supp-failurecase1}, we show a failure case where 
the source and listener are close. Accurately predicting the first-arrival spike is challenging due to its narrow ROI, which limits gradient flow. Nonetheless, our method still outperforms the baseline.

\end{document}